\documentclass{article}

\usepackage{CJKutf8}
\usepackage{longcat_style}

\usepackage[table,HTML]{xcolor}
\usepackage{adjustbox}
\usepackage[utf8]{inputenc} %
\usepackage[T1]{fontenc}    %
\usepackage{newunicodechar} %
\usepackage[normalem]{ulem} %
\usepackage{float}
\usepackage{url}
\usepackage{setspace}
\usepackage{pifont}
\usepackage{booktabs}
\usepackage{amsfonts}
\usepackage{nicefrac}
\usepackage{microtype}
\usepackage{lipsum}
\usepackage{natbib}
\usepackage{doi}
\usepackage{enumitem}
\usepackage{algorithmic}
\usepackage{array}
\usepackage{minitoc}
\usepackage{tabularx}
\usepackage{listings}
\usepackage{cancel}
\usepackage{multirow}
\usepackage{multicol}
\usepackage{makecell}

\usepackage{caption}
\usepackage{subcaption}
\usepackage{graphicx}
\usepackage[inkscapelatex=false]{svg}
\graphicspath{{figures/}}

\usepackage{amssymb}
\usepackage{amsmath} 

\usepackage{hyperref, cleveref}
\hypersetup{
    colorlinks=true,      %
    linkcolor=blue,      %
    urlcolor=blue,       %
    citecolor=blue,      %
    linkbordercolor=blue, %
    urlbordercolor=blue,
    citebordercolor=blue,
    pdfborderstyle={/S/U/W 1}, %
}

\setlist[itemize]{leftmargin=*}
\setlist[enumerate]{leftmargin=*}
\setlist[description]{leftmargin=*}

\title{LongCat-Image Technical Report}

\author{ Meituan LongCat Team \\
	\texttt{longcat-team@meituan.com} \\
}

\renewcommand{\headeright}{\raisebox{-0.2\height}{\includegraphics[height=1.8em]{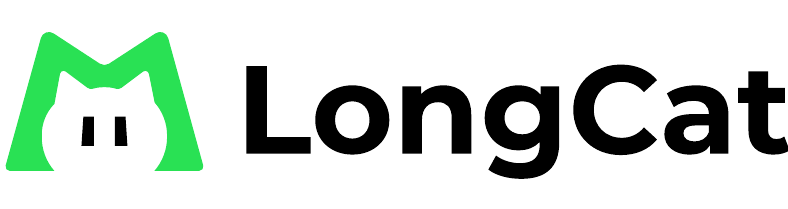}}}
\renewcommand{\shorttitle}{LongCat-Image Technical Report}

\begin{document}
\maketitle

\begin{abstract}

We introduce \textbf{LongCat-Image}, a pioneering open-source and bilingual (Chinese-English) foundation model for image generation, designed to address core challenges in multilingual text rendering, photorealism, deployment efficiency, and developer accessibility prevalent in current leading models.
1) We achieve this through rigorous data curation strategies across the pre-training, mid-training, and SFT stages, complemented by the coordinated use of curated reward models during the RL phase. This strategy establishes the model as a new state-of-the-art (SOTA), delivering superior text-rendering capabilities and remarkable photorealism, and significantly enhancing aesthetic quality.
2) Notably, it sets a new industry standard for Chinese character rendering. By supporting even complex and rare characters, it outperforms both major open-source and commercial solutions in coverage, while also achieving superior accuracy.
3) The model achieves remarkable efficiency through its compact design. With a core diffusion model of only 6B parameters, it is significantly smaller than the nearly 20B or larger Mixture-of-Experts (MoE) architectures common in the field. This ensures minimal VRAM usage and rapid inference, significantly reducing deployment costs. Beyond generation, LongCat-Image also excels in image editing, achieving SOTA results on standard benchmarks with superior editing consistency compared to other open-source works.
4) To fully empower the community, we have established the most comprehensive open-source ecosystem to date. We are releasing not only multiple model versions for text-to-image and image editing, including checkpoints after mid-training and post-training stages, but also the entire toolchain of training procedure. We believe that the leading performance, high efficiency, and openness of LongCat-Image will provide robust support for developers and researchers, collectively pushing the frontiers of multilingual visual content creation.

\vspace{10pt}

\textbf{LongCat Chat}: \href{https://longcat.ai}{https://longcat.ai} \\
\textbf{Hugging Face}: \href{https://huggingface.co/meituan-longcat/LongCat-Image}{https://huggingface.co/meituan-longcat/LongCat-Image}\\
\textbf{GitHub}: \href{https://github.com/meituan-longcat/LongCat-Image}{https://github.com/meituan-longcat/LongCat-Image} \\

\begin{figure}[!h]
    \centering
    \includegraphics[width=0.99\textwidth]{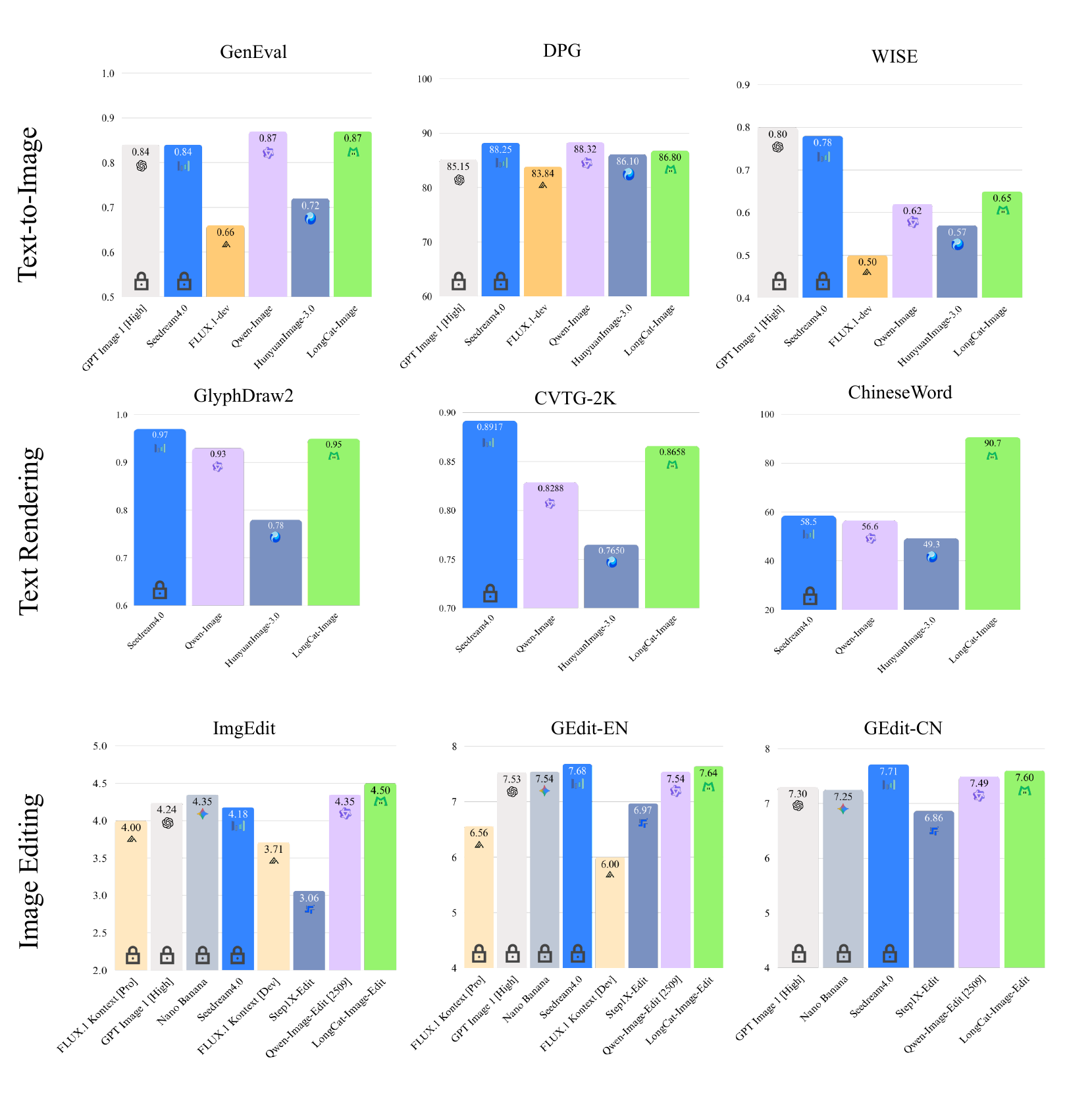}
    \caption{\textbf{LongCat-Image exhibits strong performance in general text-to-image generation, text rendering and image editing.}}
\label{fig:edit_general_compare3}
\end{figure}

\end{abstract}

\clearpage
\tableofcontents
\clearpage
\begin{figure}[!htb]
    \centering
    \includegraphics[trim={0cm 0cm 0cm 0cm}, clip, width=0.99\linewidth]{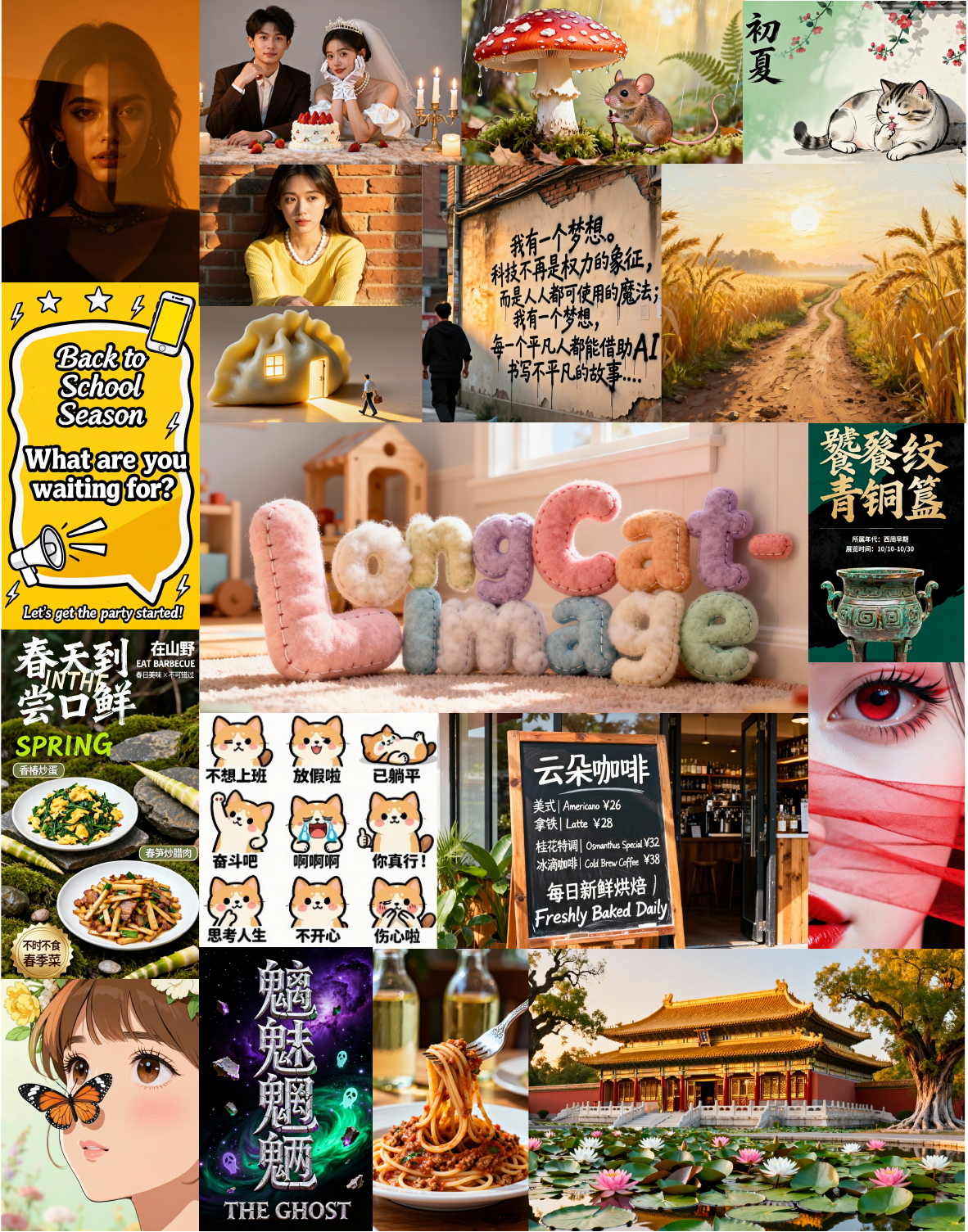}
    \caption{\textbf{High-fidelity text-to-image generation results.}}
    \label{fig:t2i_gallery}
\end{figure}

\begin{figure}[!htb]
    \centering
    \includegraphics[width=0.92\linewidth]{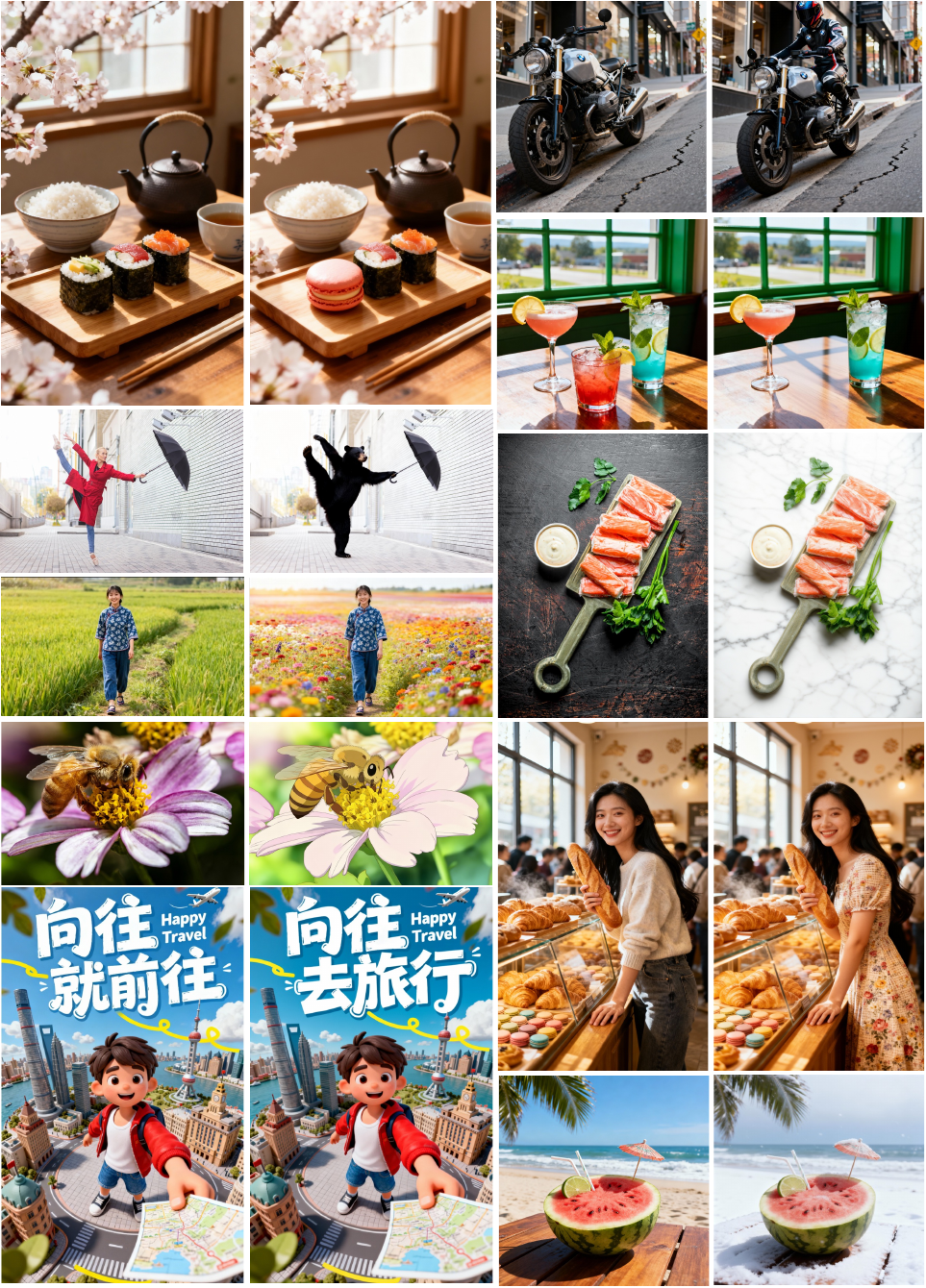}
    \caption{\textbf{Showcase of versatile capabilities in general image editing.}}
    \label{fig:edit_album1}
\end{figure}

\begin{figure}[!htb]
    \centering
    \includegraphics[width=0.97\linewidth]{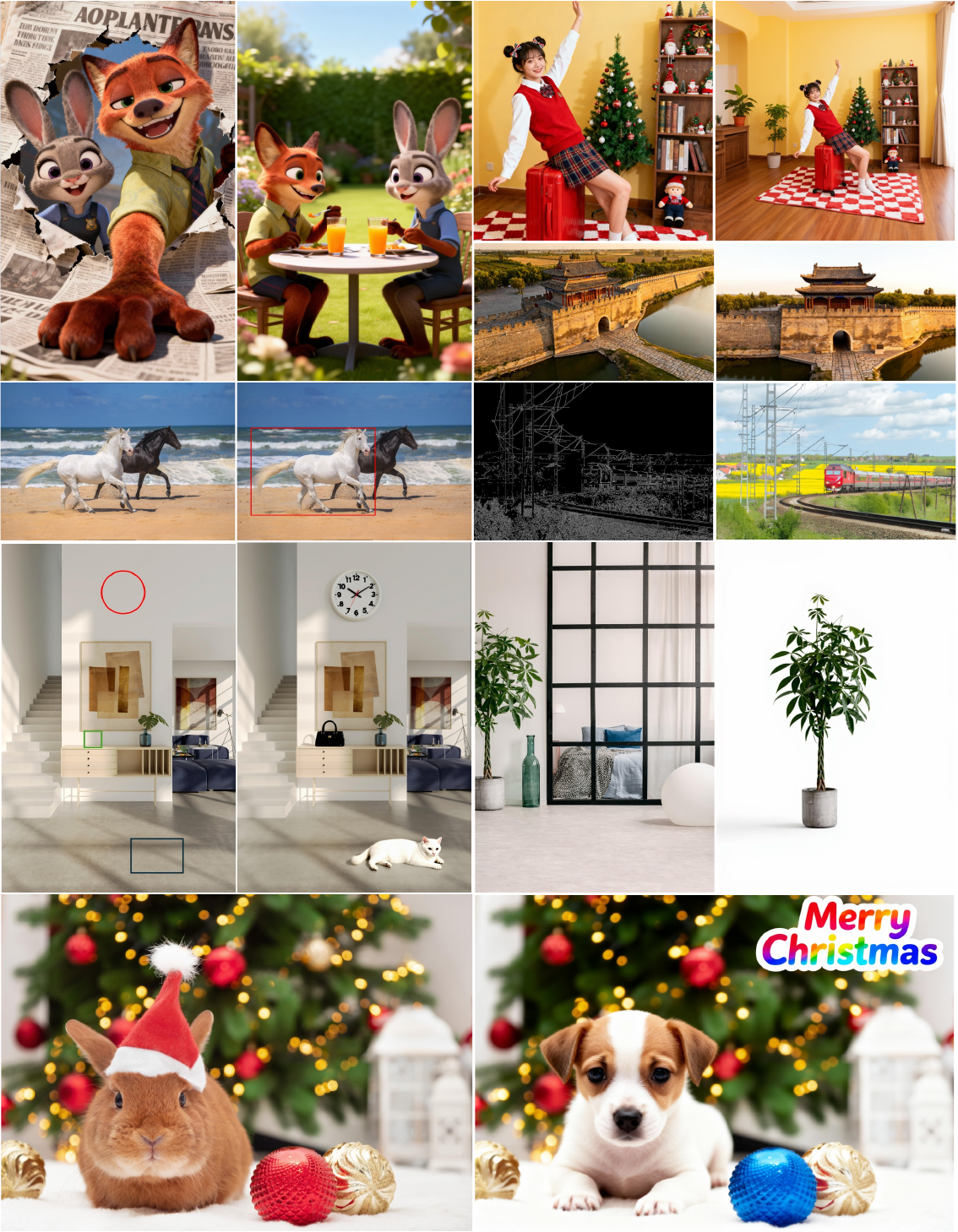}
    \caption{\textbf{Showcase on complex and comprehensive editing scenarios.} Beyond basic edits, LongCat-Image-Edit exhibits robust handling of intricate modifications and composite instructions.}
    \label{fig:edit_album2}
\end{figure}

\section{Introduction}
\label{sec:intro}

In recent years, significant advancements in Diffusion Models (DM)~\citep{ho2020ddpm,chen2023pixartalpha,black2024flux,wu2025qwenimage} have revolutionized the field of image generation, rapidly propelling the technology from academic research to widespread commercial applications, leading to the emergence of milestone products like Midjourney~\citep{midjourney2025} and Seedream~\citep{gong2025seedream2,gao2025seedream3,seedream2025seedream4}. As the technology has matured, the evaluation criteria for text-to-image (T2I) models have shifted from foundational metrics like instruction following and visual plausibility to more demanding benchmarks focusing on three core pillars: photorealism, aesthetics, and text rendering capabilities.

Concurrently, image editing has emerged as another critical domain, gaining prominence with the release of various open-source and commercial products (\textit{i.e.}, Flux.1 Kontext~\citep{batifol2025fluxkontext}, Nano Banana (Gemini-2.5-flash-image)\footnote{\url{https://aistudio.google.com/models/gemini-2-5-flash-image}}). The primary challenges in image editing currently center on two key problems: executing editing instructions with high precision and maintaining strict visual consistency between the original and edited images~\citep{wang2025image}. Although existing work~\citep{batifol2025fluxkontext,wang2025seededit3,wu2025qwenimage} has made important strides, a significant gap remains in achieving a seamless and reliable editing experience.

To address these challenges in both generation and editing, a prevailing trend has been the dramatic scaling of model parameters—from PixArt-$\alpha$~\citep{chen2023pixartalpha} at 0.6B, to Stable Diffusion3.0~\citep{esser2024sd3_0} at 8B, and further to Qwen-Image~\citep{wu2025qwenimage} at 20B and even larger Mixture-of-Experts (MoE) architectures like Hunyuan-3.0~\citep{cao2025hunyuanimage3} with 80B full parameters. The expectation has been that, similar to Large Language Models (LLMs), diffusion models would experience a breakthrough in performance through brute-force scaling. However, our observations reveal a critical issue: unrestrained parameter growth has not delivered the anticipated qualitative leap. Instead, it has led to a host of problems, including soaring computational costs, higher deployment barriers, and increased inference latency. This not only hinders the democratization of the technology but also poses challenges for open academic research.

In this context, and guided by the LongCat team's consistent design philosophy of ``Building efficient and powerful model'', we introduce \textbf{LongCat-Image}—a novel, lightweight diffusion model for image generation and editing. We contend that a more optimal equilibrium must be struck between state-of-the-art performance and efficiency in training and inference. Through systematic experimentation, we determine that a parameter scale of 6B serves as the ideal foundation for balancing capability and efficiency without compromising generative quality. Specifically, the model's core diffusion architecture employs a hybrid MM-DiT and Single-DiT structure, consistent with Flux1.dev~\citep{black2024flux}, while leveraging the Qwen2.5VL-7B~\citep{bai2025qwen2_5vl} as its text encoder to provide a unified and powerful conditional space for both generation and editing tasks.

To further enhance photorealism, we implement a systematic overhaul of our data pipeline. We observe that even a small proportion of AIGC-contaminated data can cause the model to prematurely converge to a narrow local optimum during training. While this may accelerate initial convergence, it severely limits the model's potential to achieve higher levels of realism during subsequent fine-tuning. Consequently, we rigorously exclude all AIGC data during the pre-training and intermediate training stages. In the Supervised Fine-Tuning (SFT) phase, any high-quality synthetic data introduced was meticulously hand-selected. Finally, during the Reinforcement Learning (RL) phase, we innovatively incorporate an AIGC detection model as one of the reward models, using its adversarial signal to guide the model toward generating images with the texture and fidelity of the real physical world.

To overcome the industry-wide challenge of complex Chinese text rendering, we adopt a comprehensive strategy spanning data, architecture, and training. On the data front, we utilized the SynthDoG tool~\citep{kim2022donut} to generate a large volume of text-in-image data, primarily with monotonous backgrounds, to minimize interference and improve the model's focus on learning character glyphs. Architecturally, we modify the text encoder to apply character-level encoding to the text designated for rendering in the prompt (identified by `` ''), which effectively reduces the memorization burden and enhances learning efficiency. During training, we use real-world text-in-image data in the SFT phase and introduce OCR and aesthetic reward models in the RL phase, significantly improving both the accuracy of text rendering and its natural integration with the background.

To solve the core challenge of visual consistency in image editing, we develop a meticulous training paradigm and a stringent data filtering strategy. We deliberately choose to initialize the editing model with weights from the mid-training stage of the T2I model, rather than from a highly optimized state after SFT or RL. The latter models exist in a narrowed state space, which is less conducive to learning and generalizing across diverse editing tasks. During the pre-training and SFT phases of the editing model, we employ multi-task joint training, combining editing tasks with T2I tasks. This approach effectively mitigates catastrophic forgetting of generative knowledge and enhances the model's comprehension of editing instructions. Data quality is paramount for ensuring visual consistency. We filter out samples with poor visual consistency during pre-training using a task-specific strategy and exclusively use high-consistency, human-annotated data during the SFT phase. Experiments confirm that this series of measures enables our model to achieve an exceptional standard in both instruction-following accuracy and visual consistency.

To empower the broader academic and industrial ecosystem, we are not only open-sourcing the final model but also releasing the mid-training checkpoint as a development model. Furthermore, we are providing the complete training and fine-tuning codebase, covering the entire workflow from pre-training to RL. Our goal is to foster a thoroughly open and accessible development ecosystem.

Our main contributions are five-fold:
\begin{itemize} 
\item \textbf{Exceptional Efficiency and Performance:} With only 6B parameters, LongCat-Image surpasses numerous open-source models that are several times larger across multiple benchmarks, demonstrating the immense potential of efficient model design.

\item \textbf{Remarkable Photorealism:} Through an innovative data strategy and training framework, our model achieves remarkable photorealism in generated images.

\item \textbf{Powerful Chinese Text Rendering:} The model demonstrates superior accuracy and stability in rendering common Chinese characters compared to existing SOTA open-source models and achieves industry-leading coverage of the Chinese dictionary.

\item \textbf{Superior Editing Performance}: The LongCat-Image editing model achieves state-of-the-art performance among open-source models, delivering a leading performance of instruction following and image quality, as well as superior visual consistency.

\item \textbf{Comprehensive Open-Source Ecosystem:} We provide a complete toolchain, from intermediate checkpoints to the full training code, significantly lowering the barrier for further research and development within the community.
\end{itemize}

\section{Data}
\label{sec:datac}
\begin{figure}[t]
    \centering
    \includegraphics[width=\textwidth, trim=0in 0in 0in 0in, clip]{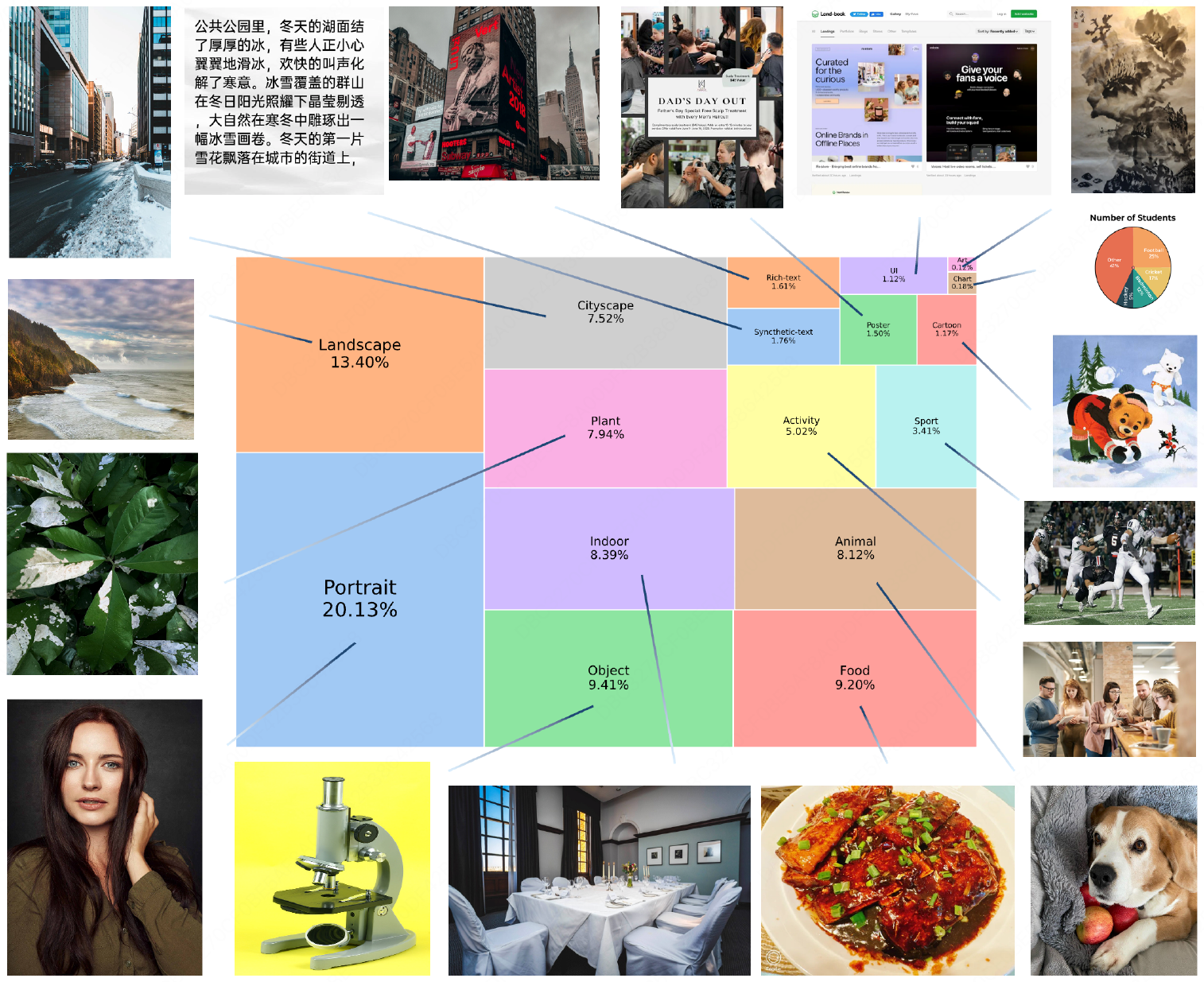}
    \caption{\textbf{Overview of training data}.}
    \label{fig:data_stat}
\end{figure}

The performance of generative models depends critically on the scale, diversity, and quality of the training corpus. Accordingly, we curate a massive dataset comprising \textbf{1.2 billion} samples. Fig.~\ref{fig:data_stat} provides a detailed statistical overview of the data composition.

\subsection{Data Curation}
\begin{figure}[t]
    \centering
    \includegraphics[width=\textwidth, trim=0cm 0cm 0cm 0cm, clip]{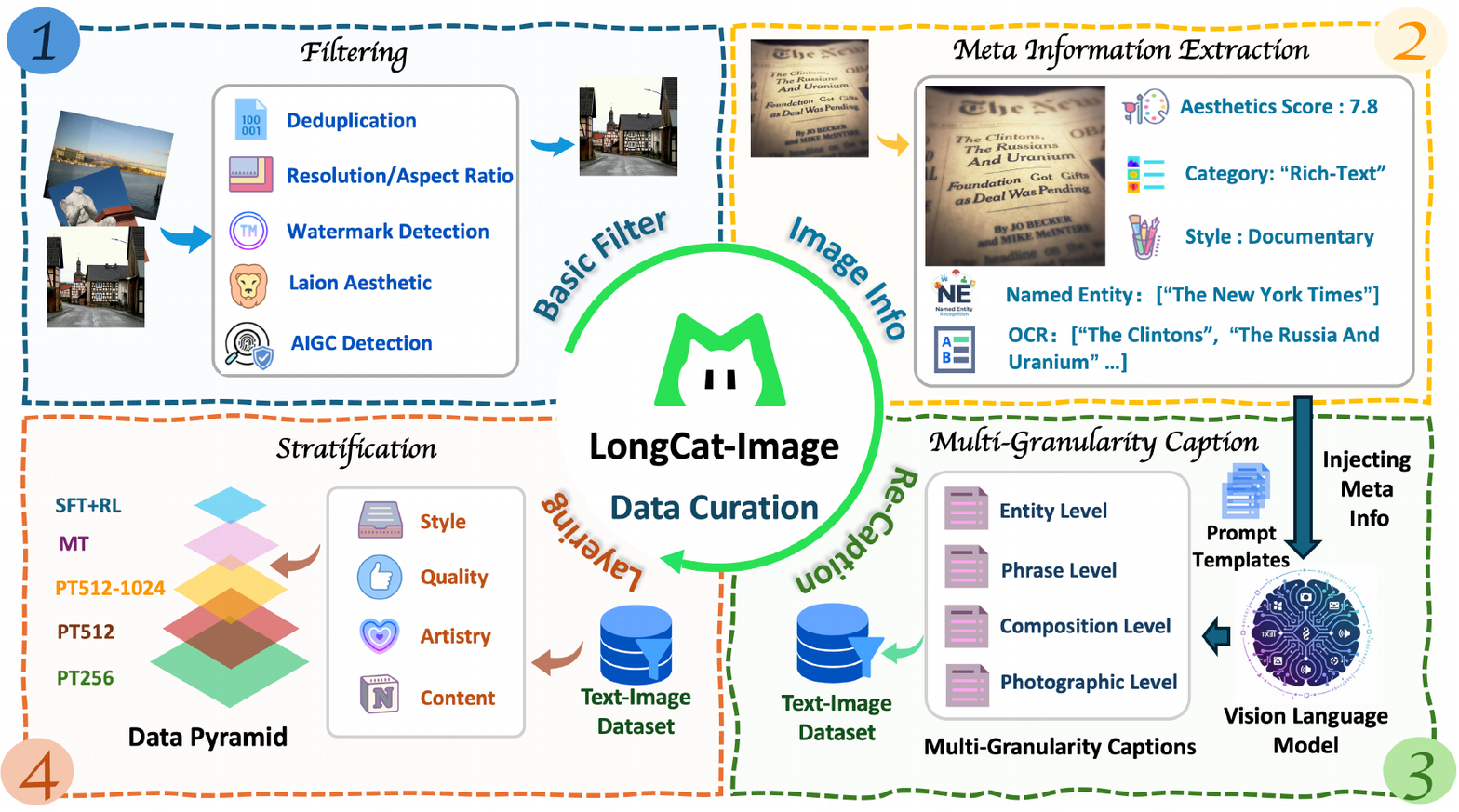}
    \caption{\textbf{Data curation pipeline.} The pipeline consists of four stages: (1) \textbf{Filtering}: Raw data undergoes deduplication and quality assessment, including watermark and AIGC detection. (2) \textbf{Meta Information Extraction}: We extract comprehensive metadata, such as aesthetic scores, named entities, and OCR text. (3) \textbf{Multi-Granularity Captioning}: Leveraging the extracted metadata and prompt templates, a VLM generates captions ranging from entity-level tags to detailed photographic descriptions. (4) \textbf{Stratification}: The dataset is stratified into a pyramid structure based on style, quality, and content to support progressive training stages.}
    \label{fig:data_curation}
\end{figure}

As illustrated in Fig.~\ref{fig:data_curation}, our data curation pipeline consists of four stages: filtering low-quality and duplicate samples, image metadata extraction, recaptioning, and data stratification for multi-stage training.

\subsubsection{Filtering}

\textbf{Deduplication.}
To address data redundancy across diverse sources, we employ a two-tiered deduplication strategy: first, MD5 hashing is used to detect exact duplicates; second, SigLIP-based similarity assessment is applied to identify and eliminate near-duplicate entries.

\textbf{Resolution \& Aspect Ratio.}
Low-resolution images and extreme aspect ratios often correlate with poor visual quality. We exclude images with a shortest edge below 384 pixels. Furthermore, only images with aspect ratios between 0.25 and 4.0 are retained to ensure a uniform and structurally coherent dataset.

\textbf{Watermark Detection.}
Watermarked images can introduce undesirable artifacts into the generated outputs. To mitigate this, we utilize a specialized watermark detector to identify and remove samples exhibiting visible watermark patterns.

\textbf{Laion Aesthetics.}
To guarantee a baseline of visual quality, each image is evaluated using the LAION-Aesthetics predictor~\citep{schuhmann2022laion}. We discard images with scores below 4.5. This threshold is empirically selected to filter out low-quality samples while preserving sufficient diversity for model training.

\textbf{AIGC Detection.}
Our experiments indicate that a small fraction of AI-generated content (AIGC) in the training data can disrupt optimization, resulting in a ``plastic'' or ``greasy'' texture in generated images. Consequently, we develop an internal AIGC detector to purge synthetic data from the corpus.

\subsubsection{Meta Infomation Extraction}

We delineate five key attributes essential for downstream processing: category, style, named entity, OCR, and aesthetics. In particular, category, style, and aesthetics are instrumental in achieving balanced data distribution and enabling hierarchical data structuring. Conversely, named entities and recognized text content play a pivotal role in enhancing the accuracy and informativeness of image captions.

\textbf{Category.} 
The category represents the semantic classification of an image according to its visual content. In our work, images are assigned to one of the following categories: \textit{portrait}, \textit{sport}, \textit{activity}, \textit{plant}, \textit{animal}, \textit{food}, \textit{object}, \textit{landscape}, \textit{cityscape}, \textit{indoor}, \textit{UI}, \textit{cartoon}, \textit{chart}, \textit{rich-text}, \textit{poster}, and \textit{synthetic text}. This categorization scheme provides a structured framework for organizing the dataset and supports subsequent processes such as content analysis and distribution balancing.

\textbf{Style.} 
Style serves as an indicator of the artistic characteristics inherent in an image. In our approach, we instruct the open-source VLM to produce a set of plausible style descriptions in the form of phrases, rather than constraining the output to a fixed set of predefined labels.

\textbf{Named Entity.} 
Named entities serve as indicators of the world knowledge encapsulated within visual content. In our work, we employ the available VLMs to identify potential celebrities, fictional characters, biological species, commercial brands, and intellectual properties depicted in the images. Subsequently, this extracted semantic information is incorporated into the subsequent image recaptioning process to enhance descriptive accuracy and contextual richness.

\textbf{OCR Text.} 
To enhance text rendering performance, an OCR model is employed to extract textual information from the images. The extracted text is subsequently processed and integrated into the corresponding image captions through a specialized handling procedure.

\begin{figure}[t]
    \centering
    \includegraphics[width=\textwidth, trim=0cm 6.5cm 0cm 6.5cm, clip]{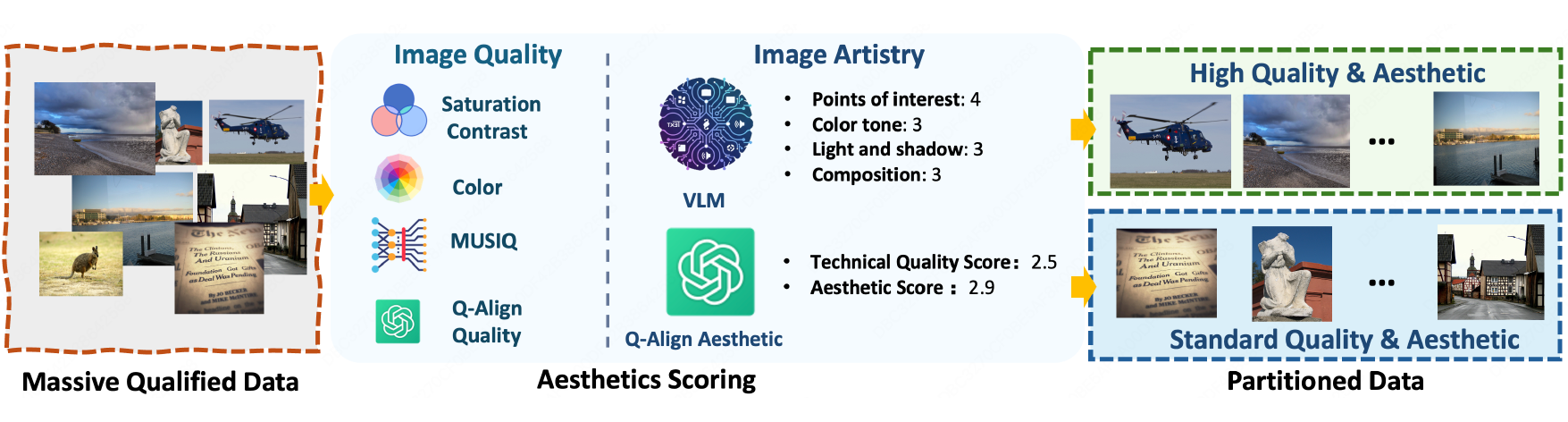}
    \caption{\textbf{Comprehensive aesthetic scoring}. The aesthetic scoring takes image quality and image artistry into account. In this context, quality denotes the signal-related attributes of the image, such as resolution, clarity, and noise levels, while artistry pertains to the perceptual appeal or visual attractiveness of the image as judged by humans or models.}
    \label{fig:hq_data}
\end{figure}

\textbf{Comprehensive Aesthetics Evaluation.} 
Quantifying aesthetics is inherently challenging due to subjectivity and the generalization limits of existing single-metric models. To address this, we decouple image evaluation into two orthogonal dimensions, namely \textit{Quality} and \textit{Artistry}, and employ an ensemble of six complementary assessment methodologies. The overall pipeline is illustrated in Fig.~\ref{fig:hq_data}.

\begin{itemize}
    \item \textit{Quality}: This dimension measures technical fidelity. We combine low-level signal statistics, including saturation, contrast, and color richness in RGB and HSV spaces, with deep reference-free assessment metrics derived from MUSIQ~\citep{ke2021musiq} and Q-Align~\citep{wu2023qalign}.
    
    \item \textit{Artistry}: This dimension evaluates photographic merit and artistic expression. We leverage VLM-based analysis to assess high-level attributes such as composition, lighting, shadow, and color tonality, complemented by the Q-Align-Aesthetics~\citep{wu2023qalign} score.
\end{itemize}

\subsubsection{Mutli-Granularity Captioning}
Image captioning with advanced Vision-Language Models (VLMs) has recently emerged as a prominent paradigm. However, there exist three critical limitations: (1) insufficient integration of world knowledge embedded within generated captions; (2) restricted diversity in caption formats; and (3) low information density resulting from verbose captions.

The lack of world knowledge often leads to inaccurate or incomplete depictions of named entities, thereby undermining the semantic fidelity of the generated images. Furthermore, captions produced by the same VLM frequently conform to similar structural patterns and lengths, limiting robustness. Finally, verbose descriptions consume valuable token space, thereby reducing the efficiency of content representation within constrained caption lengths.

To overcome the aforementioned limitations, we introduce a Multi-Granularity Captioning (MGC) framework that systematically organizes semantic abstraction into four hierarchical levels.
Specifically, the \textit{Entity Level Caption} aims to identify and describe the principal visual entities present in an image; the \textit{Phrase Level Caption} encapsulates salient visual attributes using concise linguistic expressions; the \textit{Composition Level Caption} provides an integrative interpretation that captures the overall semantic structure of the scene; and the \textit{Photographic Level Caption} offers finest-grained visual depictions, incorporating both the specific content of the image and relevant world knowledge in a succinct yet informative manner.

 \begin{figure}[b]
    \centering
    \includegraphics[width=\textwidth, trim=0cm 6cm 0cm 7cm, clip]{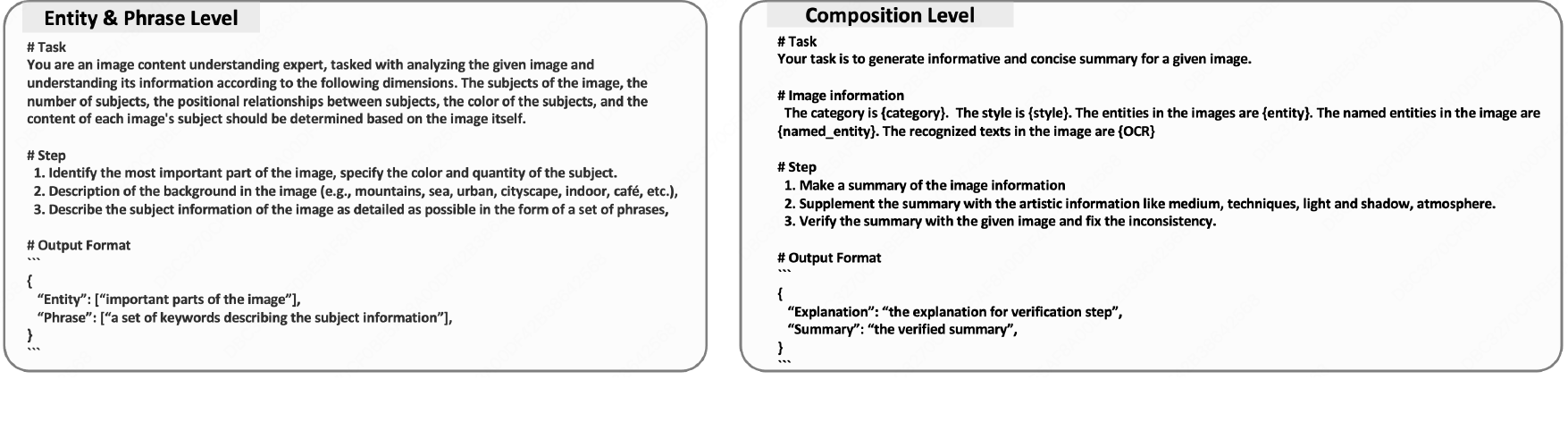}
    \caption{\textbf{Prompts for different level captions.} Entity Level and Phrase Level captions are generated concurrently using a single model, whereas composition-level captions are subsequently produced in a sequential stage utilizing a separate model.}
    \label{fig:prompt_template}
\end{figure}

At the Entity Level and Phrase Level, we employ Qwen2.5-VL~\citep{bai2025qwen2_5vl} to concurrently extract the relevant semantic information from the input image. Subsequently, at the Composition Level, we integrate the image itself, the Entity Level descriptions, and the extracted image meta-information, and feed them into InternVL2.5~\citep{chen2024internvl2_5} to generate a comprehensive, Composition Level caption. These example prompts are shown in Fig.~\ref{fig:prompt_template}.

\begin{figure}[t]
    \centering
    \includegraphics[width=\textwidth, trim=0cm 0cm 0cm 0cm, clip]{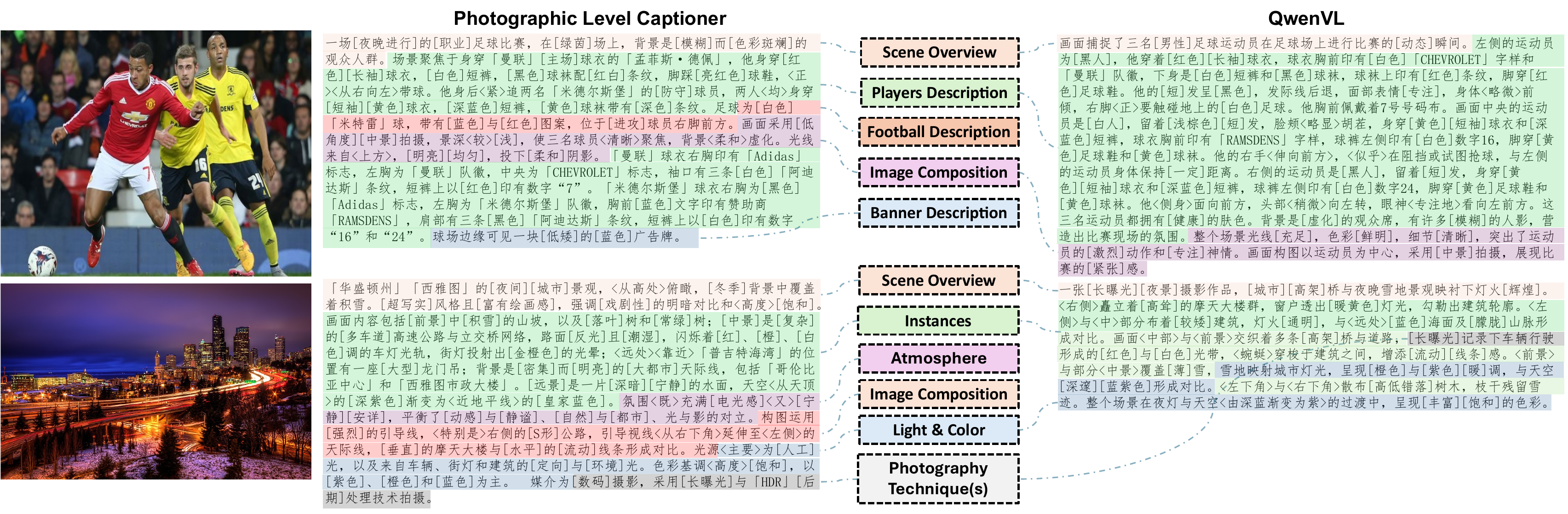}
    \caption{\textbf{Quality comparison of our Photographic Level Captioner}. This captioner produces more concise and information-dense captions compared to baseline. Different color blocks indicate different aspects of the captions.}
    \label{fig:caption}
\end{figure}

At the Photographic Level, we develop a customized captioning model, called the Photographic Captioner, based on the Qwen2.5‑VL backbone. Empirical analysis reveals that, while this open‑source backbone can produce descriptions enriched with extensive world knowledge, the output format exhibits notable inconsistencies. To mitigate this issue, we apply LoRA to fine‑tune the model, using meticulously annotated synthetic image–text pairs. This approach improves the informational density of the captions while retaining their embedded world knowledge. A qualitative comparison of caption outputs is provided in Fig.~\ref{fig:caption}.

\begin{figure}[t]
    \centering
    \includegraphics[width=\textwidth, trim=0cm 0cm 0cm 0cm, clip]{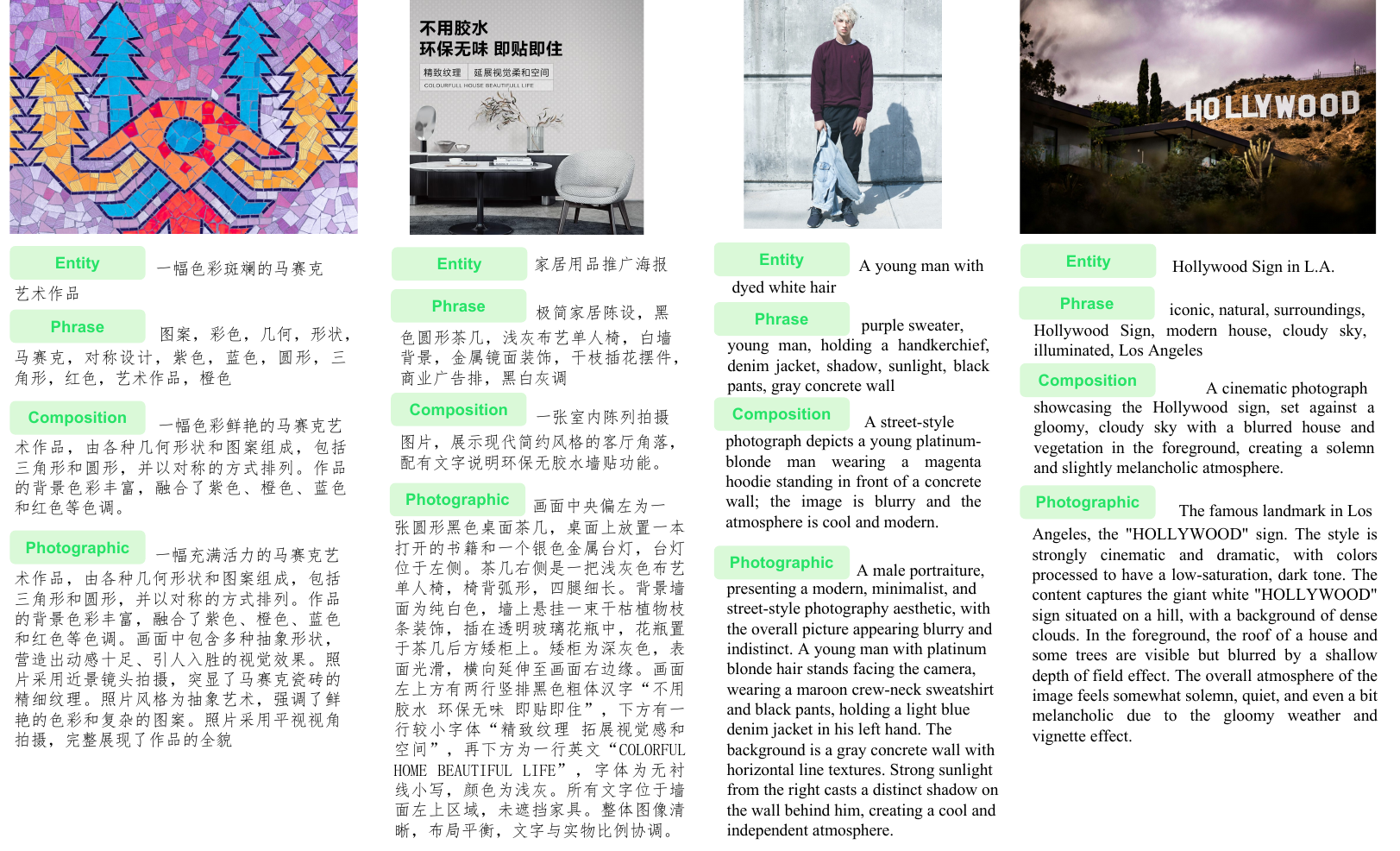}
    \caption{\textbf{Examples of Multi-Granularity Captioning.}}
    \label{fig:caption_samples}
\end{figure}

During training, we employ a weighted sampling strategy for these multi-granularity captions, prioritizing detailed descriptions to maximize information density. Specifically, the sampling probabilities for the four increasing levels of granularity are set to $[0.05, 0.1, 0.2, 0.65]$, respectively. This distribution enables the model to accommodate diverse prompt formats while robustly encoding complex world knowledge. Fig.~\ref{fig:caption_samples} illustrates representative training samples across these granularity levels.

\subsubsection{Stratification}
We stratify the training corpus into distinct subsets tailored to specific training stages, utilizing the extracted metadata—specifically image style, semantic content, and aesthetic scores.

\noindent\textbf{Pre-training.}
Empirical evidence suggests that exposing the model to a high concentration of artistic data (e.g., illustrations, cartoons, and anime) during the early pre-training stage biases the model towards learning simplified visual patterns. This tendency can compromise the model's ability to generate high-fidelity photorealistic images, effectively causing a ``collapse'' in the realistic generation subspace. Consequently, we restrict artistic data to approximately 0.5\% of the pre-training corpus, deferring the integration of broader stylistic data to the subsequent mid-training phase.

\noindent\textbf{Mid-training.}
The mid-training phase focuses on two objectives: \textit{Quality Enhancement} and \textit{Artistic Style Injection}. 
For quality enhancement, we curate a subset of high-resolution images (exceeding 1,024 pixels) from the pre-training corpus, selected for their superior sharpness, balanced composition, and high aesthetic scores. 
Simultaneously, for artistic style injection, we reintroduced the previously filtered artistic data to unlock the model's stylistic capabilities. We gradually increase the proportion of stylized data from 0.5\% to 2.5\% following a calibrated schedule. This progressive integration strategy effectively expands the model's stylistic repertoire while preserving its photorealistic foundation.

\noindent\textbf{SFT.}
In the SFT phase, we employ a mix of real and synthetic data to align the model with human aesthetic preferences. 
For real data, human experts manually curate high-fidelity samples from the Mid-training dataset, evaluating dimensions such as composition, lighting, color tonality, and emotional expression, while ensuring a balanced categorical distribution. 
Complementing this, we incorporate model-synthesized images that have undergone rigorous manual filtering to eliminate structural distortions, visual unreality, and aesthetic flaws. The strong stylistic consistency of this synthetic data facilitates the model's rapid convergence towards the manifold of human preferences.

\subsection{Data Synthesis}

\textbf{Synthetic Data Generation.} 
To address the long-tail distribution inherent in real-world datasets, we construct a specialized synthetic corpus targeting rare concepts and corner cases. Specifically, we train multiple domain-specific LoRA adapters on limited samples to capture infrequent compositional patterns and distinctive artistic styles. These adapters generate high-quality synthetic images, which are integrated into the mid-training phase at a controlled low ratio. This strategy effectively boosts performance on tail categories without compromising the diversity of the generated output space.

\begin{figure}[!htb]
    \centering
    \includegraphics[width=\textwidth, trim=0cm 0cm 0cm 0cm, clip]{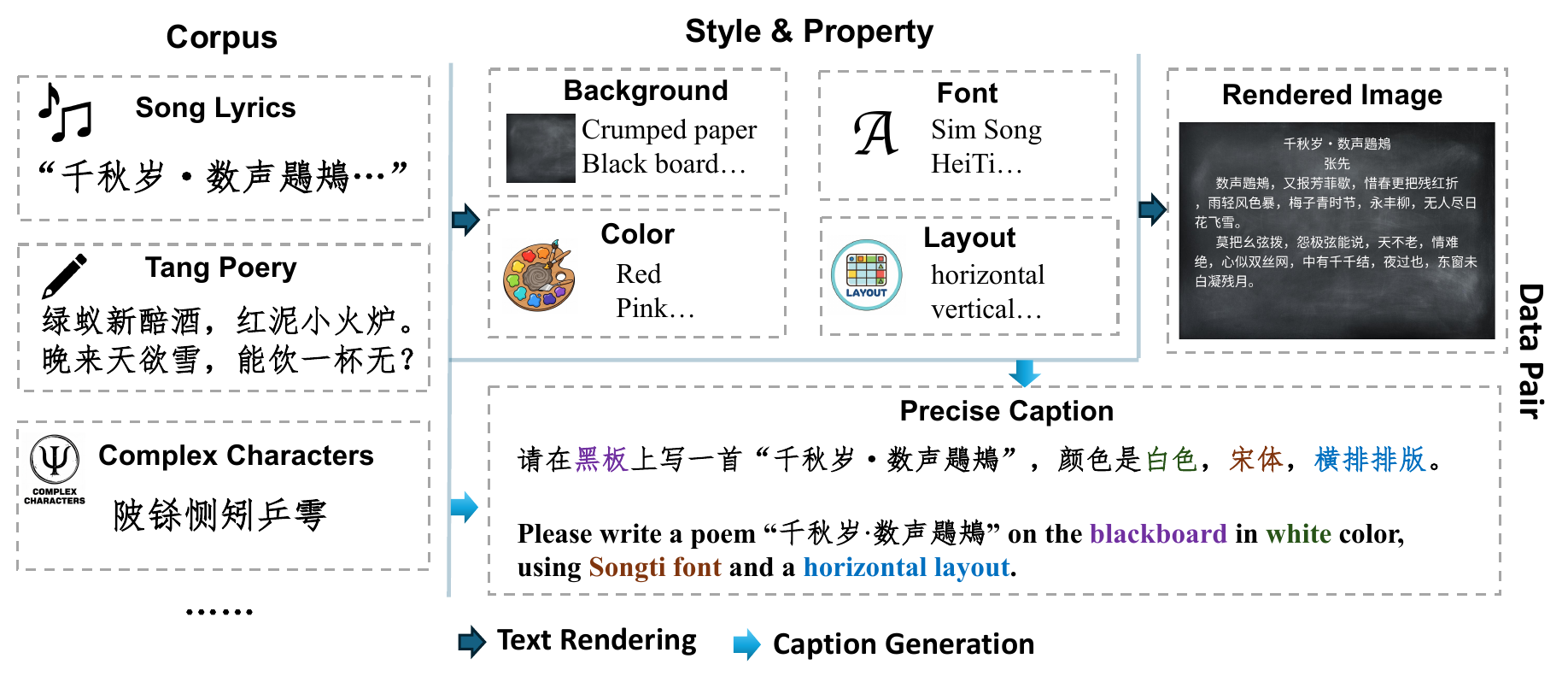}
    \caption{\textbf{Process for synthesizing text rendering data.}}
    \label{fig:syn_text}
\end{figure}

\textbf{Text Rendering.}
Empirical evidence suggests that mastering textual structures synergistically enhances a model's ability to generate other structured visual elements, thereby improving overall scene coherence. Motivated by this, we integrate synthetic text data into the pre-training corpus. As illustrated in Fig.~\ref{fig:syn_text}, our pipeline renders text from classical literature onto diverse textures, utilizing varied color palettes and fonts.

\section{Model Design}
\label{sec:model_design}

\subsection{Diffusion Model}
\begin{figure}[!htb]
    \centering
    \includegraphics[trim={0cm 0cm 0cm 0cm}, clip, width=1.0\linewidth, keepaspectratio]{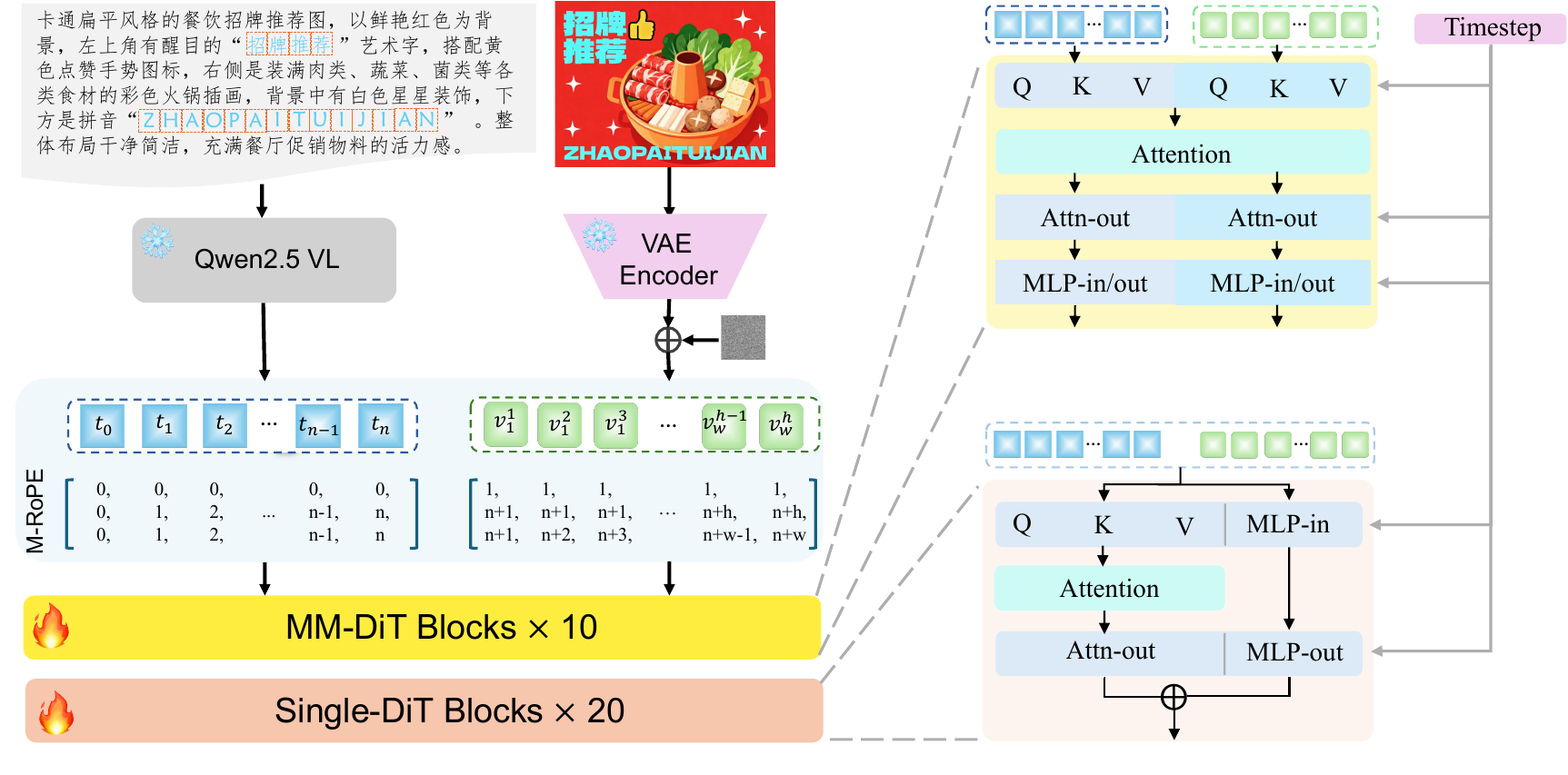}
    \caption{\textbf{Overview of model architecture.}}
    \label{fig:model_struct}
\end{figure}

We adopt the transformer architecture of FLUX.1-dev~\citep{black2024flux}, employing a double-stream attention mechanism in the initial layers and transitioning to a single-stream mechanism in the subsequent layers. To ensure parameter balance, the ratio of double-stream to single-stream blocks is maintained at approximately 1:2. The overall framework design is illustrated in Fig.~\ref{fig:model_struct}.

For the VAE component, we utilize the implementation from FLUX.1-dev. Empirical evaluations demonstrate its superior reconstruction fidelity in challenging scenarios, such as fine typography and intricate textures. Specifically, input images undergo $8\times$ spatial compression; the resulting latents are further processed via $2\times2$ token merging, yielding a final sequence length of 
$\frac{H \times W}{16\times16}$ before entering the DiT module.

\subsection{Text Encoder}
The text encoder embeds user prompts into continuous representations to condition the DiT model. While prior works~\citep{podell2023sdxl,kolors2024,li2024hunyuandit,esser2024sd3_0,black2024flux} predominantly rely on CLIP~\citep{radford2021clip} and T5~\citep{chung2024t5xxl}, recent studies~\citep{gao2025seedream3,wu2025qwenimage} have shifted toward LLMs or MLLMs to enhance multilingual compatibility, particularly for Chinese. Following this paradigm, we adopt Qwen2.5VL-7B~\citep{bai2025qwen2_5vl} as our unified text encoder. This choice ensures robust English instruction following while significantly improving Chinese processing capabilities. Furthermore, we discard the conventional injection of text embeddings into timestep embeddings for adaLN~\citep{peebles2022dit} modulation, as empirical evidence suggests negligible performance gains from this operation.

For visual text rendering, we employ a character-level tokenizer specifically for content demarcated by quotation marks. This strategy mitigates generation complexity without incurring the computational costs and memory footprint of specialized encoders (\textit{e.g.}, GlyphByT5~\citep{liu2024glyphbyt5}). Experiments demonstrate that this approach not only improves data efficiency but also accelerates convergence for text rendering tasks.

\subsection{Positional Embedding}
Positional embedding (PE) design is pivotal for handling variable aspect ratios and resolutions. While prior arts~\citep{chen2023pixartalpha,li2024hunyuandit,gong2025seedream2} rely on intricate heuristics—such as coordinate centering, frequency scaling, or interpolation—to align spatial distributions, we adopt the vanilla Multimodal Rotary Position Embedding (MRoPE)~\citep{su2024rope,wang2024qwen2vl} without modification. Our empirical observations indicate that the model possesses intrinsic adaptability to varying positional strides across different resolutions, rendering these explicit geometric constraints unnecessary. Consequently, MRoPE enables seamless generalization to unseen resolutions during pretraining without the computational or design overhead of complex adaptation strategies.

Specifically, we employ a 3D variant of MRoPE. The first dimension is designated for modality differentiation. In the text-to-image task, distinct values are assigned to distinguish tokens belonging to noise latents from those of text latents. For image editing tasks, this dimension further differentiates the latents of reference images from the aforementioned types. The remaining two dimensions encode the 2D spatial coordinates: for images, they correspond to the $(x, y)$ positions, while for text, both coordinates are set to an identical value, analogous to the behavior of 1D-RoPE. This approach not only supports flexible image generation across arbitrary aspect ratios but also facilitates seamless interaction with other modalities, such as text and reference images.

\subsection{Prompt Engineering}
To bridge the gap between the dense captions used in training and the concise, often ambiguous queries provided by users, prompt refinement is essential. While external APIs or large language models (LLMs) may offer superior rewriting capabilities, their integration often introduces deployment constraints and latency issues. To address this, we provide a default \textbf{built-in} solution that efficiently repurposes the existing condition encoder, Qwen2.5-VL. This design ensures an out-of-the-box capability for generating high-fidelity images, eliminating dependencies on external services while maintaining ease of use.
\section{Model Training}
\label{sec:training}

\begin{figure}[!tb]
    \centering
    \includegraphics[trim={0cm 0cm 0cm 0cm}, clip, width=0.65\linewidth, keepaspectratio]{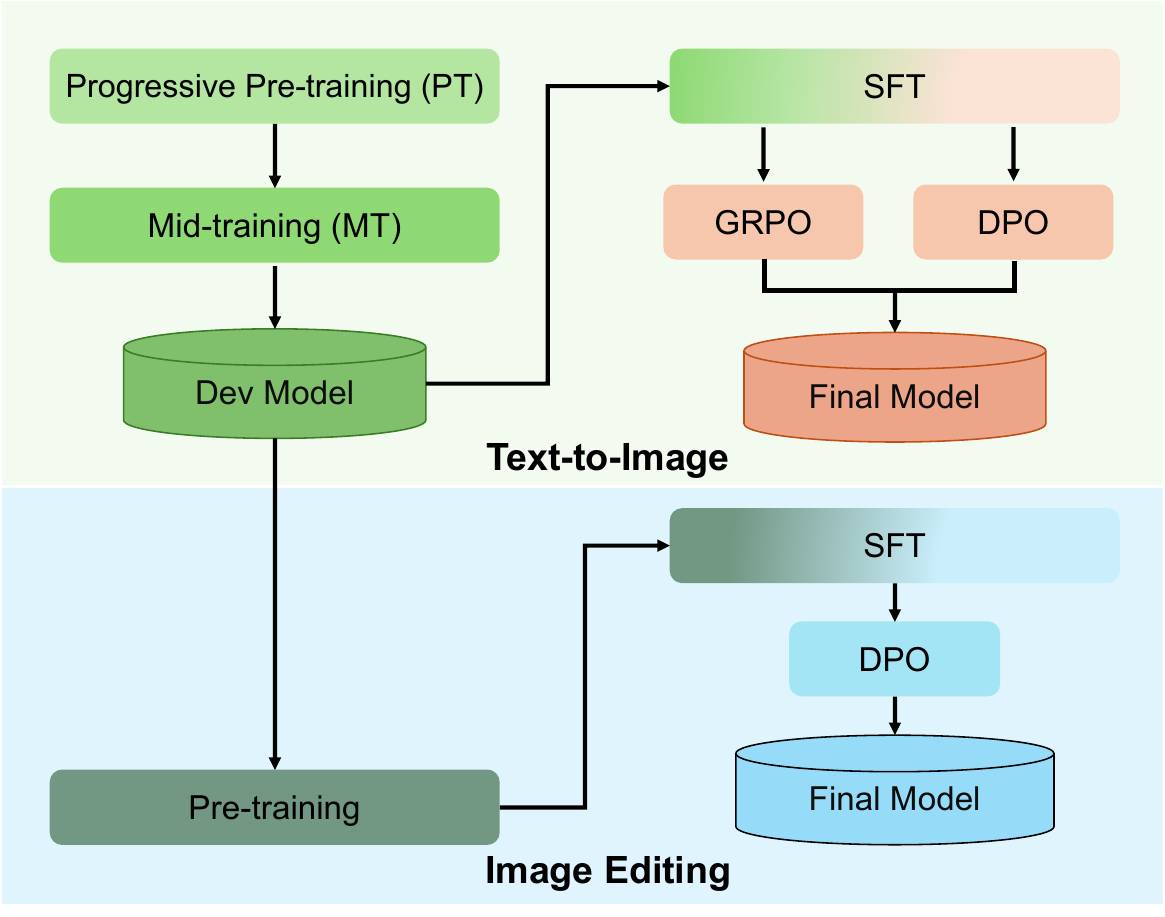}
    \caption{\textbf{Schematic overview of the multi-stage training pipeline.} The \textbf{upper panel} delineates the Text-to-Image training trajectory, progressing from progressive pre-training and mid-training to post-training alignment via SFT, GRPO, and DPO. The \textbf{lower panel} illustrates the Image Editing workflow, which initializes from the T2I development checkpoint.}
    \label{fig:train_pipeline}
\end{figure}

We establish a comprehensive multi-stage training pipeline, as illustrated in Fig.~\ref{fig:train_pipeline}, structured into three distinct phases: Pre-training, Mid-training, and Post-training.

\begin{itemize}
    \item \textbf{Pre-training}: This phase adopts a progressive multi-resolution strategy, facilitating the efficient acquisition of global semantic knowledge in early iterations while prioritizing high-frequency detail refinement in later stages.
    \item \textbf{Mid-training}: Serving as a crucial bridge between raw pre-training and alignment, this phase aims to elevate the model’s baseline generation quality. We leverage a large-scale aesthetic assessment model alongside human curation to filter a high-fidelity dataset, ensuring the underlying model possesses robust aesthetic priors.
    \item \textbf{Post-training}: The final phase focuses on alignment and stylization, comprising SFT and Reinforcement Learning (RL). In the SFT stage, we target stylized data distributions and implement a model fusion strategy to synthesize diverse stylistic capabilities. Subsequently, the RL stage incorporates advanced alignment techniques—specifically DPO and GRPO—integrating ensemble reward models to significantly enhance instruction adherence and quality.
\end{itemize}

The detailed training hyperparameters for each phase are provided in Table~\ref{tab:train-hyper}.

\begin{table}[!ht]
\caption{Progressive training recipe for LongCat-Image. }
\label{tab:train-hyper}
\vspace{3pt}
\renewcommand{\arraystretch}{1.25}
\centering
\setlength{\tabcolsep}{2pt}
\scalebox{1.0}{
    \begin{tabular}{lccccccc}
    \toprule
    & \textbf{PT $256_{px}$} & \textbf{PT $512_{px}$}  & \textbf{PT $512$-$1024_{px}$} & \textbf{MT} & \textbf{SFT}  & \textbf{DPO} & \textbf{GRPO}  \\
    \midrule
    Learning rate         & 1e{-4}   & 5e{-5} & 2e{-5} & 1e{-5} & 1e{-5} & 1e{-5} & 1e{-5}  \\
    LR scheduler       & Constant   & Constant & Constant & Constant  & Consine & Consine & Consine  \\
    Warm-up steps        & 0 & 0 & 0 & 0 & 1000 & 1000 & 0  \\
    Training steps        & 900K & 300K & 200K & 70K & 20K & 4K & 300  \\
    Global batch size        & 4608 & 4608 & 3072 & 3072 & 128 & 64 & 32  \\
    Weight decay      & \multicolumn{7}{c}{0.01} \\
    Gradient clip       & \multicolumn{7}{c}{1.0} \\
    Optimizer      & \multicolumn{7}{c}{AdamW ($\beta_1=0.9, \beta_2=0.95$)} \\
    \bottomrule
    \end{tabular}
}
\end{table}

\subsection{Pre-training}

\textbf{Progressive Mixed-Resolution Training.} We implement a progressive training curriculum commencing at $256_{px}$. To optimize training efficiency and facilitate smooth resolution adaptation, we explicitly retain an intermediate $512_{px}$ stage, avoiding the computational instability of transitioning directly from $256_{px}$ to the final phase. Subsequently, the training culminates in a dynamic stage covering a continuous resolution range between $512_{px}$ and $1024_{px}$. Throughout these phases, we employ bucket sampling to accommodate variable aspect ratios.

\textbf{Real-time Evaluation Protocol.} To facilitate dynamic strategy adjustment and convergence analysis, we implement a comprehensive monitoring system tracking validation loss, image-text alignment, aesthetic scores, and OCR-based text rendering accuracy. Empirical observations indicate that while these metrics serve as pivotal indicators during the pre-training phase, their discriminative power and sensitivity notably diminish during the mid-training and post-training stages as performance saturates.

\textbf{Dynamic Sampling of Synthetic Text Rendering Data.} The Chinese character set exhibits a distinct long-tail distribution, comprising approximately 3,000 common characters and over 5,000 rare ones that appear sparsely in natural data. To address this sparsity, we employ SynthDoG to generate a large-scale dataset (over 10 million samples) rendered on simple textures—such as paper, glass, and blackboards—with high typographic diversity (see Fig.~\ref{fig:syn_text}). While this synthetic data significantly enhances text rendering accuracy, it inevitably compromises the overall visual harmony. To mitigate this trade-off, we implement a dynamic sampling strategy based on real-time character-wise accuracy monitoring. Specifically, we increase the sampling probability for characters with high error rates, while gradually reducing synthetic exposure for well-learned characters in favor of real images. Furthermore, to prevent the model from overfitting to the simplistic synthetic domain, we completely phase out synthetic data in the final stage of pre-training.

\subsection{Mid-training}
While the pre-training phase successfully endows the model with robust global semantic priors and text-to-image mapping capabilities, the resulting visual outputs often lack high-fidelity textures and aesthetic coherence due to the inherent noise in large-scale pre-training data. The Mid-training stage is therefore designed to constrain the learned manifold, guiding the model’s distribution toward a subspace characterized by superior aesthetic quality and visual realism. This refinement serves as a critical initialization for subsequent post-training optimization.

To achieve this, we implement a rigorous data curation protocol that is significantly more stringent than that of the pre-training phase. Our pipeline integrates a hierarchical assessment system comprising advanced aesthetic scoring models, image quality estimators, and domain-specific classifiers, culminating in a human-in-the-loop verification process. This systematic filtration yields a high-fidelity corpus of millions of samples, ensuring balanced representation across diverse domains. Continued training on this curated dataset significantly elevates the model's generation quality and visual fidelity.

We designate the model derived from this stage as a foundational Developer Version. Unlike fully aligned models subjected to extensive RL, this version retains high plasticity and adaptability, avoiding the potential mode collapse or rigidity often introduced by aggressive alignment. We release this model to the community to facilitate downstream fine-tuning and further research.

\subsection{Post-training}

\subsubsection{SFT}

The primary objective of the SFT phase is to elevate the model's visual aesthetics through a rigorous data-centric approach. This involves optimizing photorealistic attributes, such as compositional integrity, lighting dynamics, and photographic techniques, while simultaneously ensuring stylistic fidelity across various artistic domains.

\textbf{High-Fidelity Data Curation}. We employ a hybrid dataset comprising hundreds of thousands of samples that blends real-world imagery with high-quality synthetic data. To prevent the degradation of realism often associated with synthetic artifacts, we enforce a strict expert verification protocol. This process ensures that only data possessing superior textural quality and aesthetic value are retained, and it strictly filters out any samples that might compromise the model's generation fidelity.

\textbf{Model Weight Averaging}. Recognizing that diverse training subsets yield models with complementary strengths, we fine-tune multiple candidate models where each is specialized in distinct visual dimensions, including illumination, portraiture, and artistic style. To synthesize these capabilities, we adopt a model parameter averaging strategy. By merging the weights of these specialized models, we effectively balance performance across multiple attributes. This fusion process significantly enhances overall robustness and stability, effectively mitigating the specific biases or deficits inherent in individual single-domain models.

\textbf{Optimization of Timestep Sampling.} Unlike the pre-training phase, which prioritizes global structural formation, SFT focuses on refining high-frequency details that typically emerge during the later stages of the diffusion process. Consequently, we transition from the Logit-Normal sampling strategy to Uniform Sampling. This adjustment ensures balanced exposure across all timesteps, specifically increasing the training weight of high-frequency denoising steps to maximize the model's capacity for learning intricate textures and fine details.

\subsubsection{RLHF}

We develop fine-grained reward models (RMs), including distortion detection, AIGC detection, human preference assessment, and OCR accuracy, to comprehensively evaluate the model's detailed capabilities. Using these RMs, we employ three distinct RL strategies: Direct Preference Optimization~(DPO)~\citep{rafailov2023dpo, wallace2024diffusiondpo}, Group Relative Policy Optimization~(GRPO)~\citep{xue2025dancegrpo}, and our proposed Monolithic Policy Optimization~(MPO). DPO excels at offline preference modeling for flow-matching models with high computational efficiency, whereas GRPO and MPO perform on-policy sampling during training with reward model evaluation. MPO fundamentally improves upon GRPO by eliminating the group-relative paradigm and its associated synchronization bottlenecks, achieving superior training efficiency and stability. To leverage the scalability advantages of offline preference learning, we conduct relatively large-scale RL with DPO and reserve on-policy methods~(MPO or GRPO) for small fine-grained RL refinement. Details of each algorithm are provided below.

\subsubsubsection{\textbf{(A) Direct Preference Optimization~(DPO)}}

\textbf{Data Construction} 1) PromptSet: A category-balanced PromptSet is constructed from real user queries of public datasets, refined through clustering and data-cleaning techniques to ensure representativeness and diversity. 2) ImagePair: We utilize diverse random initialization seeds to generate 6 candidate images for each prompt sample. An annotation team then assigns subjective quality scores ($1\text{--}5$) to each sample. To ensure clarity and effectiveness of training, we discard neutral samples (score = 3), treating high-quality ($4\text{--}5$) samples as positive and low-quality ($1\text{--}2$) samples as negative, thereby forming win-lose pairs for DPO training. This strategy ensured high confidence and preference discernment in the training data.

\paragraph{Algorithm}
We employ the DPO algorithm to mitigate common structural deficiencies in the model. Based on the SFT model, we construct a preference dataset through diversified data sampling and manual curation, optimizing the model as follows:

\begin{align}
    L(\theta) = & - \mathbb{E}_{(x_0^w, x_0^l) \sim \mathcal{D}, t \sim \mathcal{U}(0,T), x_t^w \sim q(x_t^w|x_0^w), x_t^l \sim q(x_t^l|x_0^l)} \nonumber \\
    & \log\sigma \Bigg(-\beta T \omega(\lambda_t) \Bigg( \| v^w - v_\theta(x_{t}^w,t)\|^2_2 - \| v^w - v_\text{ref}(x_{t}^w,t)\|^2_2 \nonumber \\
    & \quad - \left( \| v^l - v_\theta(x_{t}^l,t)\|^2_2 - \| v^l - v_\text{ref}(x_{t}^l,t)\|^2_2 \right) \Bigg)\Bigg) \label{eq:loss-dpo-1}.
\end{align}

We optimize the model according to Equation 2. During training, we further explore strategies including multi-round DPO iterations, gradient norm-based dirty data skipping, and KL constraints, conducting comparative analyses of different approaches' impacts on model performance. Ultimately, DPO significantly reduce the model's bad case rate and enhance the robustness of image generation.

\subsubsubsection{\textbf{(B) Group Relative Policy Optimization~(GRPO)}}

\paragraph{Algorithm}

After training with DPO, we perform further fine-grained training using GRPO following the DanceGRPO~\citep{ma2024deepcache} framework.
Given text hidden state $h$, the flow model predicts a group of $G$ images $\{x_0^i\}_{i=1}^{G}$ and the corresponding trajectory $\{x_T^i, x_{T-1}^i, ..., x_0^i\}_{i=1}^{G}$.
Within each group, the advantage function is formulated as:
\begin{equation}
    A_i = \frac{R(x_0^i, h) - \text{mean}(\{R(x_0^i, h)\}_{i=1}^G)}{\text{std}(\{R(x_0^i, h)\}_{i=1}^G)}.
\end{equation}

The training objective of GRPO is:
\begin{equation}
\begin{split}
    \mathcal{L}_{\text{GRPO}}(\theta) = \mathbb{E}_{h \sim \mathcal{D}, \{x_T^i, ..., x_0^i\}_{i=1}^{G} \sim \pi_{\theta}} \Bigg[ & \frac{1}{G}\sum_{i=1}^G \frac{1}{T}\sum_{t=0}^{T-1} \bigg( \min\big(r_t^i(\theta) A_i, \text{clip}(r_t^i(\theta), 1-\epsilon, 1+\epsilon) A_i\big)\bigg) \Bigg],
\end{split}
\end{equation}
where $r_t^i(\theta) = \frac{p_\theta(x_{t-1}^i | x_t^i, h)}{p_{\theta_{\text{old}}}(x_t^i | x_t^i, h)}$.

During trajectory sampling, we reformulate the deterministic flow-matching ODE as an SDE for effective exploration:
\begin{equation}
    dx_t = \left(v_t + \frac{\sigma_t^2}{2t}(x_t + (1-t)v_t)\right) dt + \sigma_t dw,
\end{equation}
with Euler-Maruyama discretization:
\begin{equation}
    x_{t+\Delta t} = x_t + \left[v_\theta(x_t, t, h) + \frac{\sigma_t^2}{2t}(x_t + (1-t)v_\theta(x_t, t, h))\right]\Delta t + \sigma_t\sqrt{\Delta t} \epsilon.
\end{equation}

\subsubsubsection{\textbf{(C) Monolithic Policy Optimization~(MPO)}}

\paragraph{Algorithm}

MPO employs a single-stream on-policy optimization paradigm that eliminates group-relative synchronization bottlenecks inherent in GRPO. For each prompt, MPO generates a single trajectory using a Stochastic Differential Equation~(SDE) sampler and performs one gradient update, achieving superior computational efficiency.

The generative process is governed by the SDE:
\begin{equation}
    d\mathbf{z} = \mathbf{v}_\theta(\mathbf{z}_t, c, t)dt + g(t)d\mathbf{w},
\end{equation}
where $g(t)d\mathbf{w}$ is the diffusion term enabling exploration within a single trajectory. Using Euler-Maruyama discretization:
\begin{equation}
    \mathbf{z}_{t+\Delta t} = \mathbf{z}_t + \mathbf{v}_\theta(\mathbf{z}_t, c, t)\Delta t + g_t\sqrt{\Delta t}\epsilon_t, \quad \epsilon_t \sim \mathcal{N}(0, \mathbf{I}).
\end{equation}

MPO incorporates three synergistic components for stable variance control:

\textbf{1. Gaussian Value Tracker with KL-Adaptive Forgetting:} A persistent Bayesian tracker maintains per-prompt reward estimates $\mathcal{V}(c) = \mathcal{N}(\mu_c, \sigma_c^2)$. The mean $\mu_c$ serves as a stable baseline, while the variance $\sigma_c^2$ quantifies epistemic uncertainty. Updates follow Kalman filter principles:
\begin{equation}
    \begin{aligned}
        K_t &= \frac{\sigma_{c,t-1}^2}{\sigma_{c,t-1}^2 + \sigma_{\text{obs}}^2}, \\
        \mu_{c,t} &\leftarrow \mu_{c,t-1} + K_t(r - \mu_{c,t-1}), \\
        \sigma_{c,t}^2 &\leftarrow (1-K_t)\sigma_{c,t-1}^2 + Q_t,
    \end{aligned}
\end{equation}
where $Q_t = \alpha \cdot D_{\text{KL}}(\pi_{\theta'} \| \pi_\theta)$ adaptively scales process noise with policy drift.

\textbf{2. Global Advantage Normalization:} Raw advantages $A = r - \mu_c$ are normalized using exponential moving averages:
\begin{equation}
    \tilde{A} = \frac{A - \mu_A}{\sqrt{\sigma_A^2} + \epsilon}.
\end{equation}

\textbf{3. Uncertainty-Powered Curriculum:} Prompt sampling probability follows $p(c) \propto \sigma_c + \eta/\sqrt{n_c + 1}$, prioritizing high-uncertainty prompts.

The policy update uses advantage-weighted regression:
\begin{equation}
    \mathcal{L}_{\text{MPO}}(\theta) = \mathbb{E}_{t, \mathbf{z}_t \sim \tau}\Big[\text{stop\_grad}(w_c \cdot \tilde{A}) \cdot \big\|\mathbf{v}_\theta(\mathbf{z}_t, c, t) - \mathbf{u}(\mathbf{z}_t, \mathbf{z}_0)\big\|^2\Big],
\end{equation}
where $\mathbf{u}(\cdot)$ is the target flow-matching vector field and $w_c = 1 + \gamma \cdot |r - \mu_c|/(\sigma_c + \epsilon)$.

\textbf{Training Strategy and Implementation Details.}
GRPO and MPO experiments are initialized from the same DPO base model. We optimize using the AdamW optimizer~\citep{loshchilov2017decoupled} with a constant learning rate of $5\times10^{-6}$ and a global batch size of 64. For MPO and GRPO, we employ a 12-step SDE sampler with Euler-Maruyama discretization. The diffusion coefficient $g_t$ is linearly annealed from 0.1 to 0 during training. For MPO-specific hyperparameters, we set the EMA decay for advantage normalization to $\lambda=0.99$, the curriculum balance coefficient to $\eta=1.0$, the adaptive scaling factor $\alpha=1.0$, and the surprise reweighting factor to $\gamma=0.5$.



\section{Model Performance}
\label{sec:model_performance}

\subsection{Benchmarks}

We conduct a comprehensive evaluation using a suite of established public benchmarks, including GenEval~\citep{ghosh2023geneval}, DPG-Bench~\citep{hu2024dpg}, and WISE~\citep{niu2025wise} for general generative capabilities, as well as GlyphDraw2~\citep{ma2025glyphdraw2} and CVTG-2K~\citep{du2025textcrafter} for text rendering proficiency. Furthermore, to rigorously assess Chinese character coverage, we construct a complete dictionary-based benchmark, \textit{ChineseWord}, following the protocol of Qwen-Image. Finally, to validate performance in production environments, we introduce a proprietary dataset focusing on business-critical scenarios such as poster design and natural scenes with text.

\subsubsection{Text-Image Alignment}

\paragraph{\textbf{GenEval}} evaluates the fine-grained controllability of generative models, specifically targeting attribute binding, quantitative relations, and spatial composition. As shown in Table~\ref{tab:geneval}, LongCat-Image exhibits superior performance on GenEval, demonstrating robust capabilities in handling complex compositional constraints and entity attributes.
\begin{table}[!htb]\centering
\caption{Quantitative evaluation results on GenEval.}\label{tab:geneval}
\vspace{3pt}
\renewcommand{\arraystretch}{1.25}
\resizebox{1.0\textwidth}{!}{
    \begin{tabular}{l|cccccc|c}
    \toprule
    \multirow{2}{*}{\textbf{Model}} & \textbf{Single} & \textbf{Two} & \multirow{2}{*}{\textbf{Counting}} & \multirow{2}{*}{\textbf{Colors}} & \multirow{2}{*}{\textbf{Position}} & \textbf{Attribute} & \multirow{2}{*}{\textbf{Overall$\uparrow$}} \\
    & \bf Object & \bf Object & & & & \bf Binding & \\
    \midrule
    Show-o~\citep{xie2024showo} & 0.95 & 0.52 & 0.49 & 0.82 & 0.11 & 0.28 &0.53 \\
    Emu3~\citep{wang2024emu3} & 0.98 & 0.71 & 0.34 & 0.81 & 0.17 &0.21 & 0.54 \\
    PixArt-$\alpha$~\citep{chen2023pixartalpha}         & 0.98          & 0.50        & 0.44     & 0.80    & 0.08     & 0.07              & 0.48     \\
    SD-3-Medium~\citep{esser2024sd3_0}      & 0.98          & 0.74       & 0.63     & 0.67   & 0.34     & 0.36              & 0.62     \\
    FLUX.1-dev~\citep{black2024flux}   & 0.98 & 0.81 & 0.74  & 0.79 & 0.22   & 0.45 & 0.66     \\
    SD-3.5-large~\citep{stabilityai2024sd3_5}     & 0.98          & 0.89       & 0.73     & 0.83   & 0.34     & 0.47              & 0.71     \\
    JanusFlow~\citep{ma2025janusflow} &0.97 & 0.59 & 0.45 & 0.83 & 0.53 & 0.42 & 0.63 \\
    Lumina-Image 2.0~\citep{qin2025lumina} & - & 0.87 & 0.67 & -      & - & 0.62 & 0.73     \\
    Janus-Pro-7B~\citep{chen2025januspro}     & 0.99          & 0.89       & 0.59     & 0.90    & 0.79     & 0.66              & 0.80      \\
    HiDream-I1-Full~\citep{cai2025hidream}          & 1.00             & 0.98       & 0.79     & 0.91   & 0.60      & 0.72              & 0.83     \\
    GPT Image 1 [High]~\citep{gptimage}           & 0.99          & 0.92       & 0.85     & 0.92   & 0.75     & 0.61              & 0.84     \\
    Seedream 3.0~\citep{gao2025seedream3} & 0.99 & 0.96 & 0.91 & 0.93 & 0.47 & 0.80 &0.84 \\
    Seedream 4.0~\citep{gao2025seedream3} & 0.99 & 0.92 & 0.72 & 0.91 & 0.76 & 0.74 & 0.84 \\
    Qwen-Image~\citep{wu2025qwenimage}  & 0.99 & 0.92 & 0.89 & 0.88 &  0.76 & 0.77 & \textbf{0.87} \\
    HunyuanImage-3.0~\citep{cao2025hunyuanimage3} & 1.00 & 0.92 & 0.48 & 0.82 & 0.42 & 0.63 & 0.72 \\
    \midrule
    \bf LongCat-Image & 0.99 & 0.98 & 0.86 & 0.86 & 0.75 & 0.73 & \textbf{0.87}  \\
    \bottomrule
    \end{tabular}
}
\end{table}

\begin{table}[!htb]
\centering
\caption{Quantitative evaluation results on DPG.}
\label{tab:dpg} 
\vspace{3pt}
\renewcommand{\arraystretch}{1.25}

\begin{tabular}{l|ccccc|c}
\toprule
\textbf{Model} & \textbf{Global} & \textbf{Entity} & \textbf{Attribute} & \textbf{Relation} & \textbf{Other} & \textbf{Overall}$\uparrow$ \\
\midrule
PixArt-$\alpha$~\citep{chen2023pixartalpha} & 74.97 & 79.32 & 78.60 & 82.57 & 76.96 & 71.11 \\
Lumina-Next~\citep{zhuo2024luminanext} & 82.82 & 88.65 & 86.44 & 80.53 & 81.82 & 74.63 \\
SDXL~\citep{podell2023sdxl} & 83.27 & 82.43 & 80.91 & 86.76 & 80.41 & 74.65 \\
Playground v2.5~\citep{Playground} & 83.06 & 82.59 & 81.20 & 84.08 & 83.50 & 75.47 \\
Hunyuan-DiT~\citep{li2024hunyuandit} & 84.59 & 80.59 & 88.01 & 74.36 & 86.41 & 78.87 \\
Janus~\citep{wu2025janus} & 82.33 & 87.38 & 87.70 & 85.46 & 86.41 & 79.68 \\
PixArt-$\Sigma$~\citep{chen2024pixartsigma} & 86.89 & 82.89 & 88.94 & 86.59 & 87.68 & 80.54 \\
Emu3-Gen~\citep{wang2024emu3} & 85.21 & 86.68 & 86.84 & 90.22 & 83.15 & 80.60 \\
Janus-Pro-1B~\citep{chen2025januspro} & 87.58 & 88.63 & 88.17 & 88.98 & 88.30 & 82.63 \\
DALL-E 3~\citep{openai2023dalle3} & 90.97 & 89.61 & 88.39 & 90.58 & 89.83 & 83.50 \\
FLUX.1-dev~\citep{black2024flux} & 74.35 & 90.00 & 88.96 & 90.87 & 88.33 & 83.84 \\
SD-3-Medium~\citep{esser2024sd3_0} & 87.90 & 91.01 & 88.83 & 80.70 & 88.68 & 84.08 \\
Janus-Pro-7B~\citep{ma2025janusflow} & 86.90 & 88.90 & 89.40 & 89.32 & 89.48 & 84.19 \\
HiDream-I1-Full~\citep{cai2025hidream} & 76.44 & 90.22 & 89.48 & 93.74 & 91.83 & 85.89 \\
Lumina-Image 2.0~\citep{qin2025lumina} & - & 91.97 & 90.20 & \textbf{94.85} & - & 87.20 \\
Seedream 3.0~\citep{gao2025seedream3} & \textbf{94.31} & \textbf{92.65} & 91.36 & 92.78 & 88.24 & 88.27 \\
GPT Image 1 [High]~\citep{gptimage} & 88.89 & 88.94 & 89.84 & 92.63 & 90.96 & 85.15 \\
Seedream 4.0~\citep{seedream2025seedream4} & 94.10 & 92.28 & 92.75 & 93.67 & 92.77 & 88.25 \\
Qwen-Image~\cite{wu2025qwenimage} & 91.32 & 91.56 & \textbf{92.02} & 94.31 & \textbf{92.73} & \textbf{88.32} \\
HunyuanImage-3.0~\citep{cao2025hunyuanimage3} & 92.12 & 92.53 & 89.13 & 92.13 & 91.92 & 86.10 \\
\midrule
\bf LongCat-Image & 89.10 & 92.54 & 92.00 & 93.28 & 87.50 & 86.80 \\
\bottomrule
\end{tabular}
\end{table}

\paragraph{\textbf{DPG-Bench}} comprises 1,065 dense and structurally complex prompts designed to challenge the semantic alignment of text-to-image models. As presented in Table~\ref{tab:dpg}, LongCat-Image achieves competitive alignment performance, ranking closely behind leading models such as Qwen-Image and Seed4.0, thereby validating its proficiency in interpreting verbose captions.

\paragraph{\textbf{WISE}} comprises 1,000 curated prompts aimed at rigorously testing semantic comprehension and world knowledge. During evaluation, we leverage the off-the-shelf text encoder for intrinsic prompt enhancement. Results indicate that LongCat-Image achieves state-of-the-art (SOTA) scores among open-source diffusion models, underscoring its robust reasoning capabilities and responsiveness to semantic nuances. Table~\ref{tab:WiScore} details these findings.
\begin{table*}[!htb]
\centering
\caption{Quantitative evaluation results of world knowledge reasoning on WISE.}
\label{tab:WiScore}
\vspace{3pt}
\renewcommand{\arraystretch}{1.25}

\begin{tabular}{l|cccccc|c} 
\toprule
Model \rule{0pt}{0.25em} & Cultural & Time & Space & Biology & Physics & Chemistry & \textbf{Overall} \rule[-0.3em]{0pt}{1.5em} \\

\midrule
\multicolumn{8}{c}{\textbf{Unified Models}} \\
\midrule
GPT4o~\citep{gptimage} & \textbf{0.81} &\textbf{0.71} & \textbf{0.89} & \textbf{0.83} & \textbf{0.79 }& \textbf{0.74} & \textbf{0.80}\\
MetaQuery-XL~\citep{pan2025metaquery} & 0.56 & 0.55 & 0.62 & 0.49 & 0.63 & 0.41 & 0.55\\
Liquid~\citep{wu2025liquid} &0.34	&0.45	&0.48	&0.41	&0.45	&0.27&	0.39\\
Emu3~\citep{wang2024emu3} & 0.34& 0.45 &0.48 & 0.41 &0.45 &0.27 & 0.39 \\
Janus-1.3B~\citep{wu2025janus} &0.16 &0.26 &0.35 & 0.28 &0.30 & 0.14& 0.23\\
JanusFlow~\citep{ma2025janusflow} &0.13 &0.26 &0.28 & 0.20& 0.19&0.11 & 0.18\\
Janus-Pro-1B~\citep{chen2025januspro} & 0.20& 0.28&0.45 & 0.24 & 0.32& 0.16& 0.26\\
Janus-Pro-7B~\citep{chen2025januspro} & 0.30& 0.37& 0.49& 0.36&0.42 &0.26 & 0.35 \\
Orthus-7B-instruct~\citep{kou2024orthus} &0.23 &0.31 &0.38 &0.28 & 0.31&0.20 & 0.27\\
Show-o-512~\citep{xie2024showo} & 0.28 &0.40 &0.48 & 0.30&0.46 &0.30 & 0.35\\
\midrule

\multicolumn{8}{c}{\textbf{Text-to-Image Models}} \\
\midrule
FLUX.1-dev~\citep{black2024flux} &0.48 & 0.58 &0.62  &0.42 &0.51 & 0.35 &  0.50 \\
FLUX.1-schnell~\citep{black2024flux} &0.39 &0.44 &0.50 & 0.31&0.44 &0.26 & 0.40 \\
PixArt-$\alpha$~\citep{chen2023pixartalpha} &0.45 & 0.50& 0.48 & 0.49 &0.56 &0.34 & 0.47\\
Playground-v2.5~\citep{Playground} & 0.49  &0.58 & 0.55&0.43 & 0.48&0.33 & 0.49 \\
SD-3-medium~\citep{esser2024sd3_0} &0.42 & 0.44 &0.48 &0.39 &0.47 &0.29 & 0.42 \\
SD-3.5-medium~\citep{stabilityai2024sd3_5} &0.43 & 0.50 &0.52 &0.41 &0.53 &0.33 & 0.45 \\
SD-3.5-large~\citep{stabilityai2024sd3_5} & 0.44 &0.50 &0.58 & 0.44&0.52 &0.31 & 0.46 \\
Seedream 4.0~\citep{seedream2025seedream4} & 0.78 & 0.73 & 0.85 & 0.79 & 0.84 & 0.67 & \textbf{0.78} \\
Qwen-Image~\citep{wu2025qwenimage} & 0.62 & 0.63 &0.77 & 0.57&0.75 &0.40 & 0.62 \\
HunyuanImage-3.0~\citep{cao2025hunyuanimage3} & 0.58 & 0.57 & 0.70 & 0.56 & 0.63 & 0.31 & 0.57 \\
\textbf{LongCat-Image} & 0.66 & 0.61 & 0.72 & 0.66 & 0.72 & 0.49 & \underline{0.65} \\
\bottomrule
\end{tabular}
\end{table*}

\subsubsection{Text Rendering}

\paragraph{\textbf{GlyphDraw2}} assesses text generation across two distinct subsets. The \textit{Poster-Set} (200 prompts) evaluates bilingual generation in design contexts, while the \textit{Complex-Set} challenges models with random combinations drawn from a pool of 2,000 frequent Chinese characters to test coverage. As illustrated in Table~\ref{tab:glydraw2}, LongCat-Image excels particularly in the Complex-Set, highlighting its robustness in rendering intricate character structures.

\begin{table}[!htb]
\centering
\caption{Quantitative evaluation results of GlyphDraw2.}
\label{tab:glydraw2}
\vspace{3pt}
\renewcommand{\arraystretch}{1.25}

\begin{tabular}{l|ccccc}
\toprule
Model & Complex-en & Complex-zh & Poster-en & Poster-zh & Avg\\ \hline
Seedream 4.0~\citep{seedream2025seedream4} & 0.99 & 0.91  &  0.99  & 0.99  & \textbf{0.97} \\ 
Qwen-Image~\citep{wu2025qwenimage}  & 0.90 &  0.87 & 0.95  & 0.98  & 0.93 \\
HunyuanImage-3.0~\citep{cao2025hunyuanimage3}    & 0.47  &  0.85 & 0.90  &  0.90 & 0.78  \\ 
\textbf{LongCat-Image}  & 0.94  & 0.92  &  0.95 &  0.99 & \underline{0.95} \\ 
\bottomrule
\end{tabular}
\end{table}

\paragraph{\textbf{CVTG-2K}} focuses on English text rendering across diverse real-world scenarios, including street views, advertisements, and memes. Each prompt features multi-region layouts (2 to 5 regions) to test spatial text placement. LongCat-Image attains SOTA performance on this benchmark (see Table~\ref{tab:glydraw2}), demonstrating exceptional effectiveness in complex, multi-turn English text rendering tasks.

\begin{table}[!ht]
\centering
\caption{Quantitative evaluation results of CVTG-2K.}
\label{tab:cvtg}
\vspace{3pt}
\renewcommand{\arraystretch}{1.25}
\resizebox{0.95\linewidth}{!}{
\begin{tabular}{l|ccccc|c|c}
\toprule
\multirow{2}{*}{Model} &  \multicolumn{5}{c|}{\bf Word Accuracy$\uparrow$} & \multirow{2}{*}{\bf NED$\uparrow$} & \multirow{2}{*}{\bf CLIPScore$\uparrow$} \\
\cmidrule{2-6} & 2 regions & 3 regions & 4 regions & 5 regions & average & & \\
\midrule
Seedream 4.0~\citep{seedream2025seedream4} & 0.8898 & 0.9147 & 0.8991  & 0.8873 & \textbf{0.8917} & \textbf{0.9507} & 0.7853 \\ 
Qwen-Image~\citep{wu2025qwenimage}   & 0.8370 & 0.8364 & 0.8313  & 0.8158 & 0.8288 & 0.9297 & \underline{0.8059}\\ 
HunyuanImage-3.0~\citep{cao2025hunyuanimage3}   & 0.8300 & 0.7635 & 0.7384  & 0.7279 & 0.7650 & 0.8765 & \textbf{0.8121} \\ 
\textbf{LongCat-Image}   & 0.9129 & 0.8737 & 0.8557  & 0.8310 & \underline{0.8658} & \underline{0.9361} & 0.7859\\ 
\bottomrule
\end{tabular}
}
\end{table}

\begin{CJK}{UTF8}{gbsn}
\paragraph{\textbf{ChineseWord}} To evaluate the full spectrum of Chinese character rendering, especially for long-tail characters, we constructed a comprehensive benchmark comprising 8,105 prompts based on the \textit{General Standard Chinese Characters Table}\footnote{\url{http://www.moe.gov.cn/jyb_sjzl/ziliao/A19/201306/t20130601_186002.html}}, aligning with the protocol of Qwen-Image. Each character is embedded in a standardized template (e.g., ``On the blackboard, the word `华' is written in purple Song font.''). We employ PPOCRv5~\citep{cui2025paddleocr} for objective accuracy quantification, as existing MLLMs often struggle to recognize rare characters. Results in Table~\ref{tab:chineseword} demonstrate that LongCat-Image outperforms all existing models by a significant margin. However, we acknowledge that while exhibiting dominance in single-character rendering, the model experiences a noticeable decline in stability when generating multi-character sequences, primarily due to the insufficient scale of real-world textual training data. In future work, we aim to address this by rigorously expanding our text-rich dataset collection to enhance robustness in complex multi-character generation tasks.
\end{CJK}
\begin{table}[!tb]
\centering
\caption{Quantitative evaluation results of ChineseWord.}
\label{tab:chineseword}
\vspace{3pt}
\renewcommand{\arraystretch}{1.25}

\begin{tabular*}{0.75\linewidth}{l|@{\extracolsep{\fill}}cccc}
\toprule
Model & L1 & L2 & L3 & Overall\\ 
\midrule
Seedream 4.0~\citep{seedream2025seedream4} & 94.8 & 41.2 & 2.3  &  \underline{58.5} \\ 
Qwen-Image~\citep{wu2025qwenimage}   & 92.5 & 37.1 & 6.1 & 56.6  \\ 
HunyuanImage-3.0~\citep{cao2025hunyuanimage3}  & 83.5  & 31.3 & 4.1 &  49.3  \\ 
\textbf{LongCat-Image}   & 98.7 & 90.8 & 70.3 & \textbf{90.7}  \\ 
\bottomrule
\end{tabular*}
\end{table}

\paragraph{\textbf{Poster\&SceneBench}} To bridge the gap between academic benchmarks and industrial applications, we curate a dataset of 500 prompts covering both poster typography and natural scene text. Unlike flat poster layouts, the latter specifically evaluates the model's capability to seamlessly integrate text into real-world environments (\textit{e.g.}, signage on textured surfaces or shop fronts with complex lighting). As indicated in Table~\ref{tab:inner_text}, LongCat-Image delivers SOTA-level performance, proving its reliability and effectiveness in these practical operational contexts.
\begin{table}[!tb]
\centering
\caption{Quantitative evaluation results of internal poster and scene text scenarios.}
\label{tab:inner_text}
\vspace{3pt}
\renewcommand{\arraystretch}{1.25}

\begin{tabular*}{0.75\linewidth}{l|@{\extracolsep{\fill}}ccc}
\toprule
Model & Poster & Real Scene & Avg\\
\midrule
Seedream 4.0~\citep{seedream2025seedream4} & 93.2 & 90.0 & \textbf{91.6}  \\ 
Qwen-Image~\citep{wu2025qwenimage}   & 88.7 & 89.6 & 89.2 \\ 
HunyuanImage-3.0~\citep{cao2025hunyuanimage3}   & 89.0 & 85.1 & 87.1 \\ 
\textbf{LongCat-Image}   & 92.0 & 91.0 & \underline{91.5} \\ 
\bottomrule
\end{tabular*}
\end{table}

\subsection{Human Evaluation}
To assess perceptual quality, we adopt the Mean Opinion Score (MOS) protocol, focusing on four distinct dimensions: text-image alignment, visual plausibility, visual realism, and aesthetics.
\begin{itemize}

\item \textbf{Text-Image Alignment} measures the semantic fidelity of the generated image to the input prompt. It evaluates the accurate depiction of key elements, including entities, attributes, spatial relationships, and stylistic constraints.

\item \textbf{Plausibility} examines the image's adherence to physical coherence and logical consistency. This metric penalizes anatomical distortions, unnatural proportions, and spatial anomalies that violate real-world physics.

\item \textbf{Realism} assesses the degree of photorealism and texture fidelity. It serves as a proxy for the ``Turing test'' of image generation, determining how indistinguishable the synthesized output is from authentic photography, free from ``AI-generated'' artifacts.

\item \textbf{Aesthetics} evaluates the artistic quality and perceptual appeal, considering factors such as composition, lighting, color harmony, and overall visual impact.

\end{itemize}

\paragraph{\textbf{Evaluation dataset.}} To ensure a rigorous and unbiased assessment, we constructed a diverse dataset of 400 prompts tailored for black-box evaluation. This corpus spans a broad spectrum of difficulty, ranging from fundamental entity depiction to intricate scene synthesis. It comprehensively covers challenging generative dimensions, including multi-entity interactions, spatial layouts, dynamic actions, artistic creativity, text rendering, and world knowledge reasoning.

\begin{figure}[!h]
    \centering
    \includegraphics[width=1.0\linewidth]{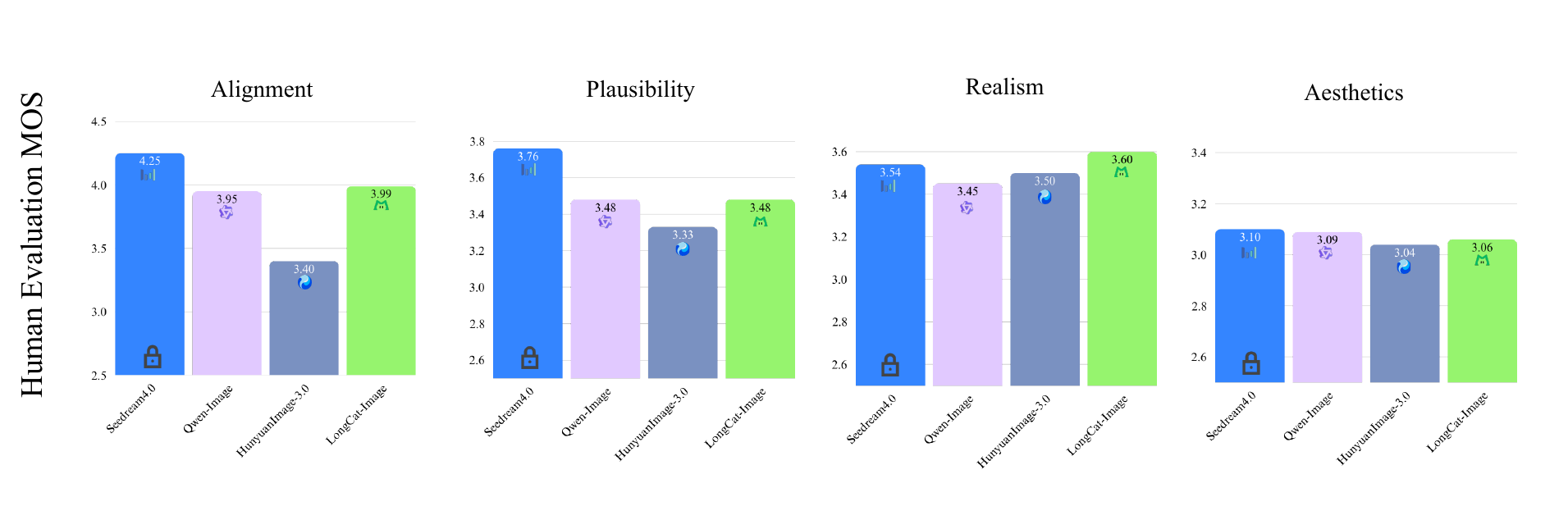}
    \caption{\textbf{Comparison of human evaluation MOS.}}
    \label{fig:human_mos}
\end{figure}

\paragraph{\textbf{Result Analysis.}} As illustrated in Fig.~\ref{fig:human_mos}, LongCat-Image demonstrates comprehensive superiority over HunyuanImage 3.0 across all metrics. Compared to Qwen-Image, our model achieves parity in both Text-Image Alignment and Visual Plausibility. Notably, LongCat-Image excels in Visual Realism, outperforming Qwen-Image and even exhibiting a slight advantage over the commercial baseline, Seedream 4.0. While there remains a marginal gap in Visual Aesthetics compared to Qwen-Image, the overall human evaluation indicates that LongCat-Image delivers performance comparable to SOTA open-source models.

\subsection{Qualitative Results}

\begin{figure}[p]
    \centering
    \includegraphics[width=1.0\linewidth]{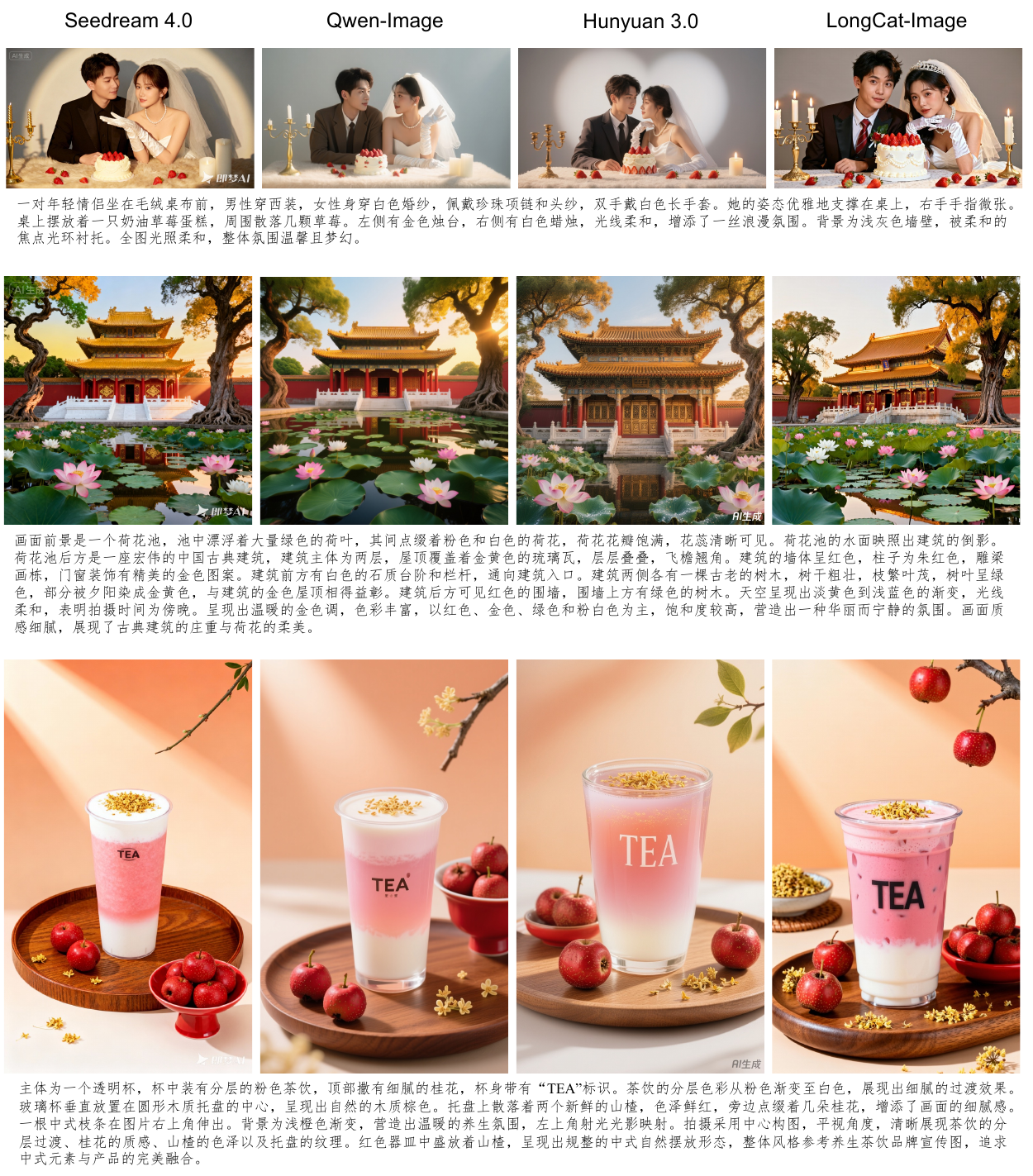}
    \caption{\textbf{Comparison of overall capability in image generation.}}
    \label{fig:compare_genernal}
\end{figure}

\begin{figure}[p]
    \centering
    \includegraphics[width=1.0\linewidth]{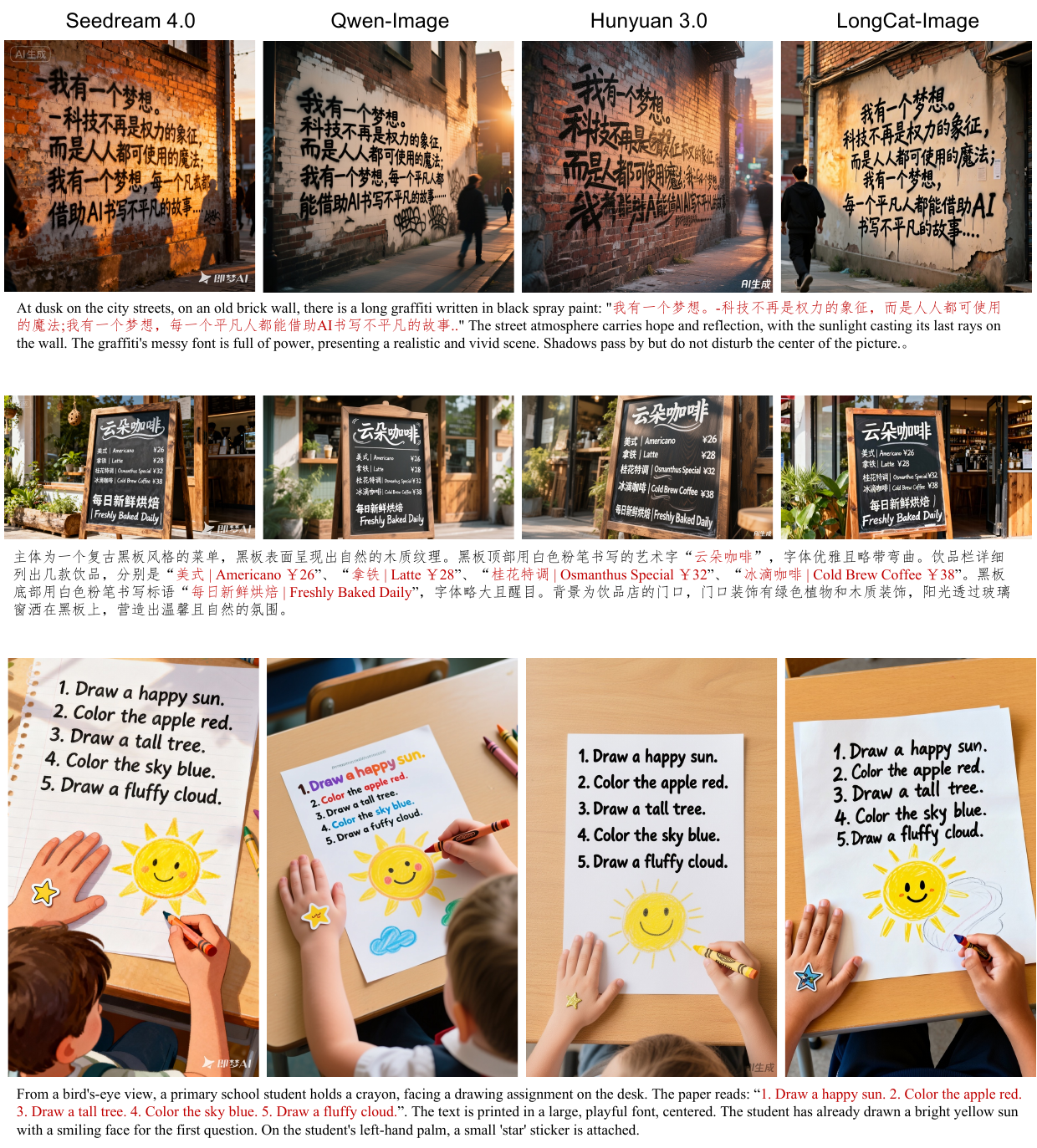}
    \caption{\textbf{Comparison of text rendering capability in image generation.}}
    \label{fig:compare_text_scene}
\end{figure}

\begin{figure}[p]
    \centering
    \includegraphics[width=1.0\linewidth]{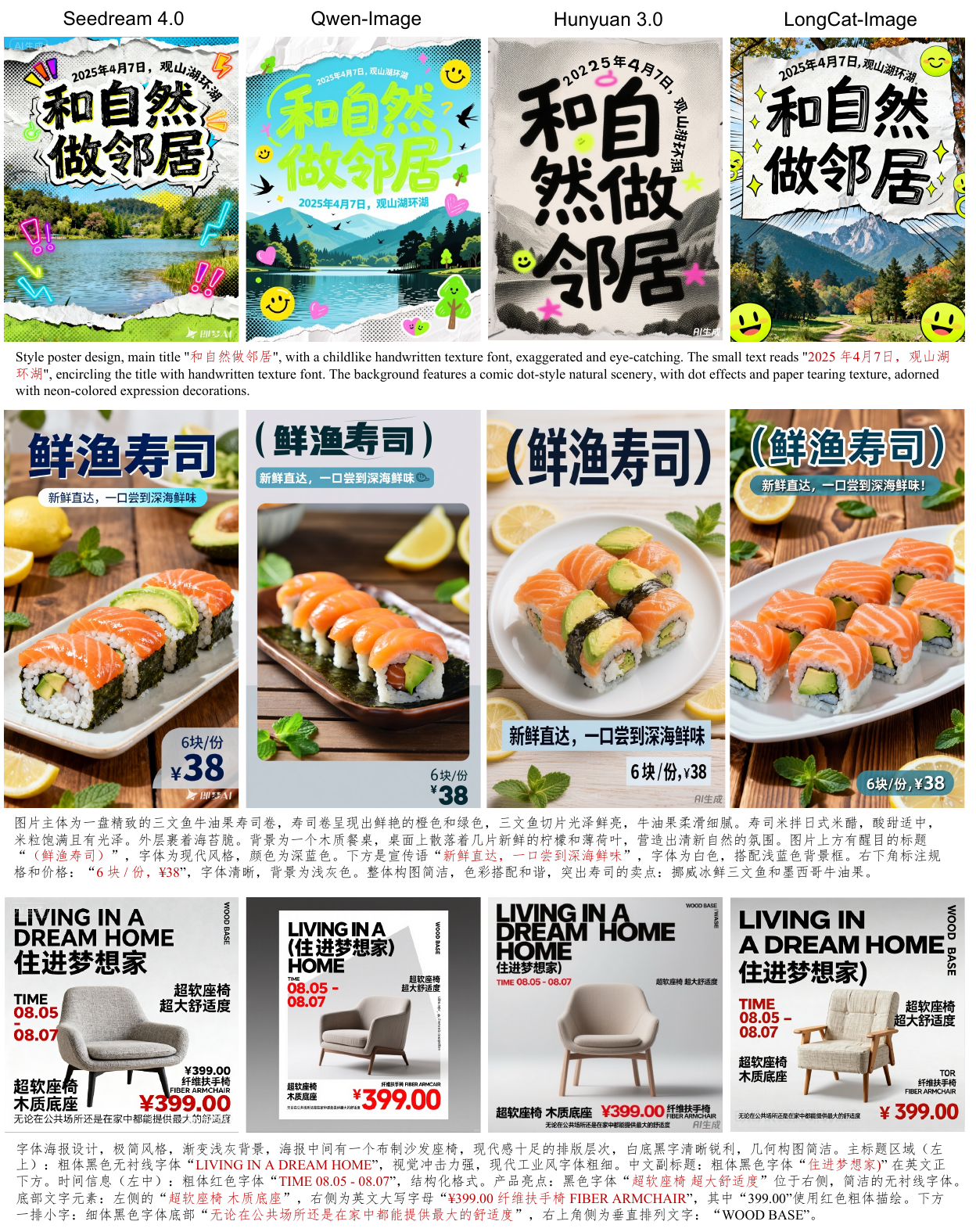}
    \caption{\textbf{Comparison of text rendering capability in image generation.}}
    \label{fig:compare_text_poster}
\end{figure}

\begin{figure}[p]
    \centering
    \includegraphics[width=1.0\linewidth]{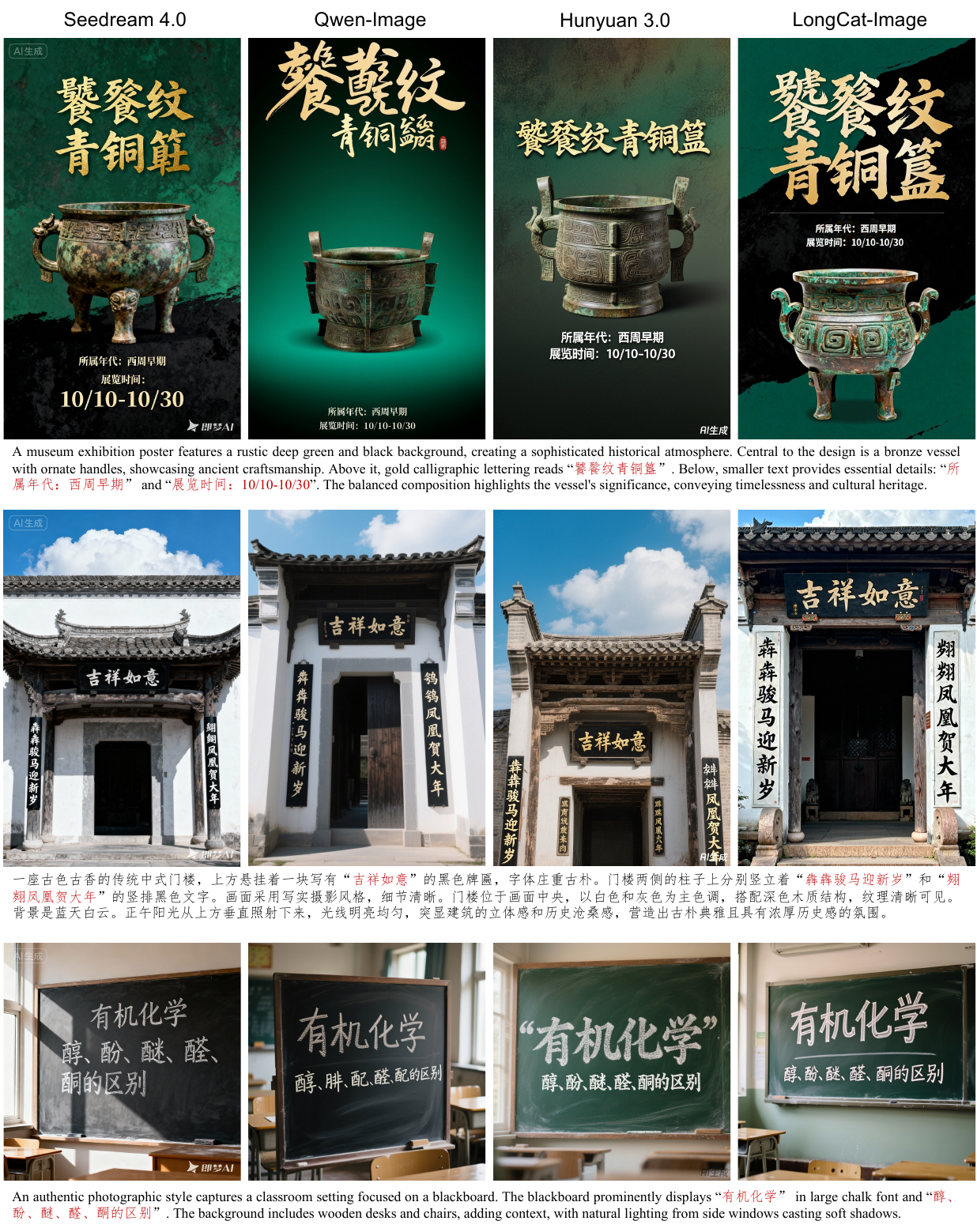}
    \caption{\textbf{Comparison of text rendering capability in image generation.}}
    \label{fig:compare_text_hard}
\end{figure}

In Fig.~\ref{fig:compare_genernal},~\ref{fig:compare_text_scene},~\ref{fig:compare_text_poster},~\ref{fig:compare_text_hard}, we present qualitative comparisons between our LongCat-Image and leading SOTA baselines. Visual inspection confirms that our model exhibits robust performance across critical dimensions, including text-image alignment, visual plausibility, realism, and aesthetic quality. Furthermore, it demonstrates exceptional proficiency in text rendering tasks.

\section{Image Editing}\label{sec:image_editing}
Extending a Text-to-Image foundation model for image editing is a well-established paradigm, effectively validated by prior works~\citep{batifol2025fluxkontext, wang2025seededit3,wu2025qwenimage}. In this section, we detail the adaptation of LongCat-Image into \textbf{LongCat-Image-Edit}, which achieves SOTA performance among open-source models.

\subsection{Data Curation}
\begin{figure}[!htb]
    \centering
    \includegraphics[width=1.0\linewidth]{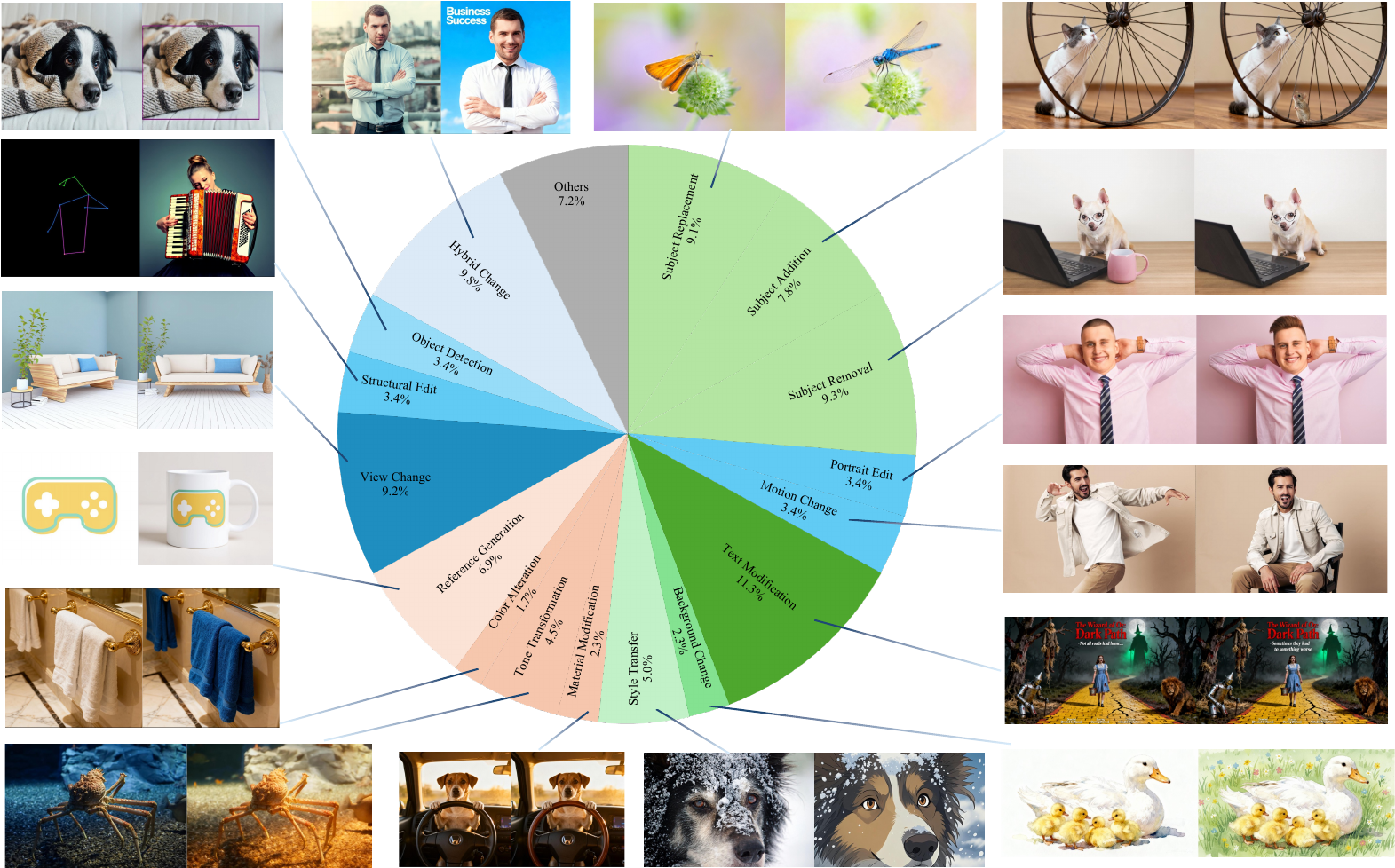}
    \caption{\textbf{Overview of image editing pre-training data.}}
    \label{fig:edit_pretraing_dataset}
\end{figure}

Distinct from standard text-to-image pre-training, image editing necessitates source-target image pairs. We curate a comprehensive training set from diverse sources, including open-source datasets, synthetic pipelines, video sequences, and interleaved web corpora. To enhance the model's instruction-following capabilities, which range from simple descriptions to complex reasoning, we apply extensive instruction rewriting strategies to maximize linguistic diversity. Fig.~\ref{fig:edit_pretraing_dataset} illustrates the proportional distribution of editing tasks within our training set.

\subsubsection{Open-Source Datasets}
Given the high cost of annotating editing pairs, we prioritize leveraging high-quality open-source repositories, specifically OmniEdit~\citep{wei2024omniedit}, OmniGen2~\citep{wu2025omnigen2}, and NHREdit~\citep{kuprashevich2025nohumansrequired}. We implement a rigorous data cleaning pipeline on these datasets and rewrite the original instructions. This process efficiently yields high-fidelity training pairs tailored for diverse editing tasks.

\subsubsection{Synthesized Data}
Specialized expert models demonstrate exceptional performance in specific common editing tasks. Leveraging their capabilities, we synthesize high-quality training pairs for tasks such as object manipulation, style transfer, background alteration, and reference-based generation. For each task, we establish a dedicated pipeline where MLLMs craft editing instructions, and the corresponding expert models generate the target images. Complementarily, we employ traditional image processing algorithms to construct data for low-level adjustments like filter transformations and lighting changes. Furthermore, we incorporate human-in-the-loop verification for critical samples to ensure semantic alignment and visual fidelity. This approach allows us to accumulate a substantial volume of high-quality data.

\subsubsection{Video Frames}
Synthetic methods often struggle with complex structural changes, such as human pose and perspective, frequently introducing artifacts. To bridge this gap, we leverage video sequences that naturally capture realistic temporal transitions. Our pipeline employs multimodal models to identify target objects and optical flow estimation to quantify changes between frames. We extract keyframe pairs that exhibit significant yet coherent variations and automatically annotate them with editing instructions. A subset of these pairs undergoes manual verification to guarantee accuracy.

\subsubsection{Interleaved Corpus}
While the aforementioned data sources effectively cover standard editing categories, they often lack coverage for long-tail scenarios. To bridge this gap and significantly enrich the diversity of editing instructions, we exploit web-scale interleaved corpora. By mining large-scale image-text sequences with inherent semantic correlations, we extract implicit editing signals from naturally occurring data. These raw pairs undergo rigorous filtering and multimodal-assisted instruction rewriting to ensure their suitability for training. However, mining valid training samples from such massive unstructured corpora is an extremely resource-intensive endeavor. Consequently, the scale of data we have curated to date remains limited. We firmly believe that this represents a critical direction for long-term data engineering and encourage broader community participation to further explore this valuable frontier.

\subsubsection{Instruction Rewriting}
Since synthetic instructions often diverge from real-world user prompts and complex reasoning benchmarks, we employ GPT-4o~\citep{hurst2024gpt} to enhance instruction diversity. We implement a one-to-many strategy, associating each editing pair with multiple rewritten variants, including natural language paraphrases and compound commands. This approach aligns training data with diverse inference scenarios, a benefit subsequently validated by our experimental results.

\begin{figure}[htb!]
    \centering
    \includegraphics[trim={0cm 0cm 0cm 0cm}, clip, width=0.85\linewidth, keepaspectratio]{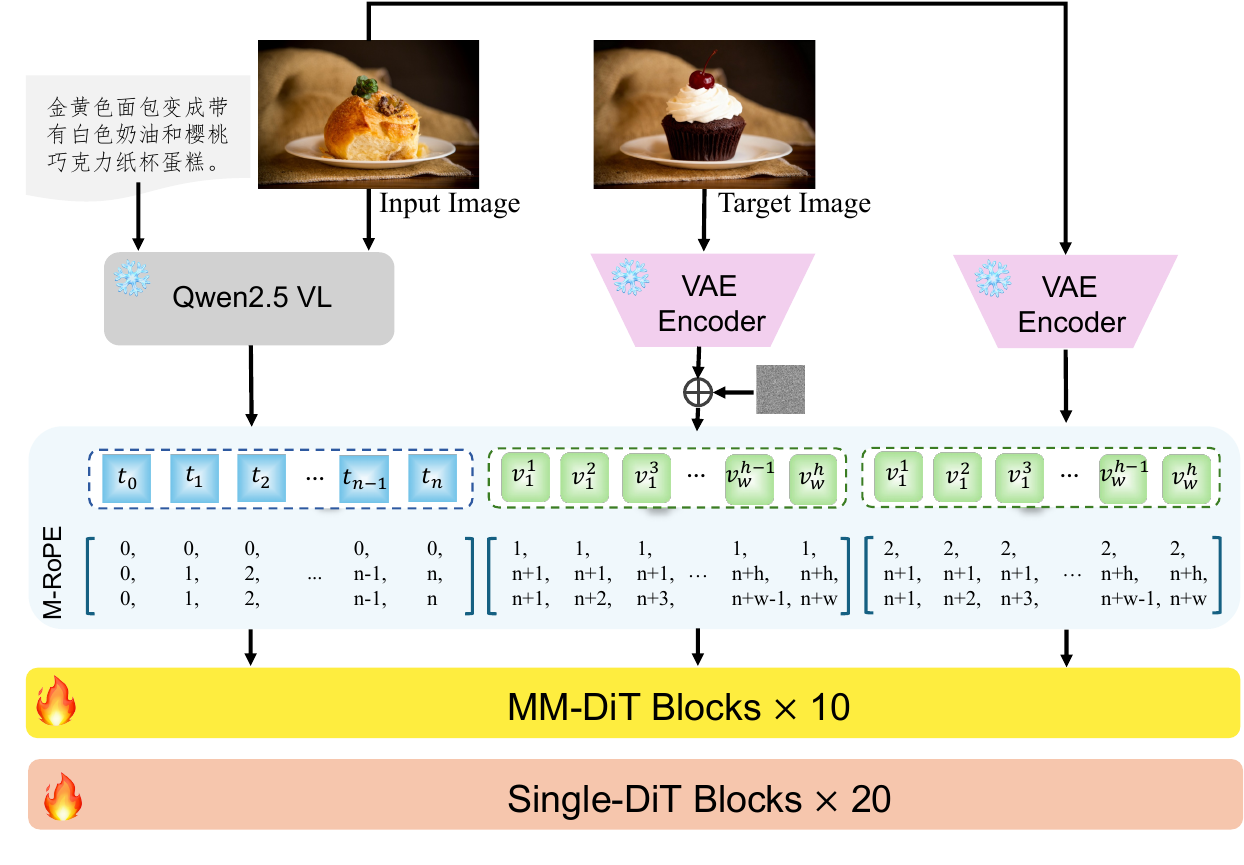}
    \caption{\textbf{Overview of LongCat-Image-Edit model architecture.}}
    \label{fig:model_struct_edit}
\end{figure}

\subsection{Model Design}
Building upon the base architecture and drawing inspiration from prior works~\citep{wang2025seededit3,wu2025qwenimage}, we introduce an image conditioning branch via modifications to VAE features, 3D RoPE embeddings, and token sequencing. Specifically, reference images are encoded into VAE latents and distinguished from noised latents by manipulating the first dimension of the 3D RoPE embeddings, while preserving spatial alignment in the remaining dimensions. These reference tokens are then concatenated with noised latents along the sequence dimension to serve as input for the diffusion visual stream. Furthermore, we feed both the source image and instructions into the multimodal encoder (\textit{i.e.}, Qwen2.5-VL). To differentiate editing tasks from standard text-to-image generation, we implement a distinct system prompt during this feature extraction process. The overall schematic of the model architecture is illustrated in Fig.~\ref{fig:model_struct_edit}.

\subsection{Model Training}
As illustrated in Fig.~\ref{fig:train_pipeline}, the training framework comprises three progressive stages: Pre-training, SFT, and DPO. This multi-stage curriculum is designed to systematically enhance the resolution and visual fidelity of the generated images.

\subsubsection{Pre-training}
We initialize the model using a mid-training T2I checkpoint, as its unconstrained parameter space offers superior plasticity compared to post-trained models. Our training follows a multi-scale strategy: we begin at 512$\times$512 resolution with massive, noisy datasets for rapid convergence, then progress to 1024$\times$1024 with high-quality data to refine details. Simultaneously, we adopt a joint training strategy, mixing editing data with T2I mid-training data at a balanced batch ratio. Since experiments confirm this approach improves both semantic understanding and image quality, we retain it for the following SFT stage. Furthermore, to enhance instruction generalization, we associate each sample with 3-5 candidate prompts (in Chinese and English) and randomly select one during each training iteration.

\subsubsection{SFT}
During the SFT stage, we curate a high-fidelity dataset comprising hundreds of thousands of samples from real photographs, professional manual retouches, and synthetic sources. To guarantee generation stability, we implement a rigorous human-in-the-loop filtering protocol, specifically targeting the structural alignment between source and edited images. Our experiments reveal a high sensitivity to data quality: even a marginal relaxation of these alignment standards leads to a precipitous drop in the model's ability to maintain consistency. Consequently, we enforce the strictest criteria to ensure precise content preservation. Furthermore, by jointly training this strictly filtered corpus with high-quality T2I SFT data, we achieve significant improvements in both instruction adherence and aesthetic quality.

\subsubsection{DPO}

To further align the model with human aesthetic standards and mitigate persistent structural artifacts, we employ Direct Preference Optimization (DPO) following the SFT stage. We construct a high-quality preference dataset via diverse sampling and rigorous manual annotation, optimizing the diffusion DPO objective defined as:

\begin{align}
    L(\theta) = & - \mathbb{E}_{(I_\text{src}^w, P^w, I_\text{src}^l, P^l) \sim \mathcal{D}, \; t \sim \mathcal{U}(0,T), \; x_t^w \sim q(x_t^w|I_\text{src}^w, P^w), \; x_t^l \sim q(x_t^l|I_\text{src}^l, P^l)} \nonumber \\
    & \log\sigma \Bigg(-\beta T \omega(\lambda_t) \Bigg( \| v^w - v_\theta(x_{t}^w, I_\text{src}^w, P^w, t)\|^2_2 - \| v^w - v_\text{ref}(x_{t}^w, I_\text{src}^w, P^w, t)\|^2_2 \nonumber \\
    & \quad - \left( \| v^l - v_\theta(x_{t}^l, I_\text{src}^l, P^l, t)\|^2_2 - \| v^l - v_\text{ref}(x_{t}^l, I_\text{src}^l, P^l, t)\|^2_2 \right) \Bigg)\Bigg) \label{eq:loss-dpo-image-edit}.
\end{align}

\textbf{Data Construction.} The preference dataset is curated in two steps.
(1) Prompt Curation: We synthesize category-balanced image-instruction pairs leveraging both Multimodal LLMs and real-world user queries. These inputs undergo clustering and filtration to ensure semantic diversity and representativeness. (2) Preference Annotation: For each unique prompt, we generate five candidate outputs using distinct random seeds. A professional annotation team evaluates these candidates to identify the most successful edit (winner) and the failed instances (losers). This rigorous selection process ensures high-confidence preference signals, forming robust win-lose pairs for training.

\textbf{Training Strategy.} We optimize the model using Eq.~\eqref{eq:loss-dpo-image-edit}. To stabilize training and prevent reward hacking, we incorporate advanced strategies such as gradient-based outlier rejection (to skip noisy data) and KL divergence constraints. Comparative analysis confirms that these techniques are crucial for performance gains. Ultimately, DPO significantly reduces the failure rate (\textit{e.g.}, structural artifacts) and enhances the overall robustness of the image generation.

\subsection{Discussion}
Initially, we aim to unify T2I and image editing into a single model to leverage potential task synergies and minimize deployment costs. However, experiments reveal a critical data quality mismatch: the heavy reliance on synthetic data during editing pre-training noticeably degrades the photorealism of T2I generation compared to models trained solely on real data. Consequently, we decided to separate the models to ensure optimal performance for each task. We emphasize that this is a data-driven issue, not an architectural flaw. We believe that by substituting synthetic datasets with large-scale interleaved corpora, future iterations can successfully merge these capabilities into a unified model without sacrificing generation quality.

\subsection{Model Performance}

In this section, we comprehensively evaluate our model across three quantitative benchmarks: CEdit-Bench (Ours)\footnote{https://huggingface.co/datasets/meituan-longcat/CEdit-Bench}, GEdit-Bench~\citep{liu2025step1x}, and ImgEdit-Bench~\citep{ye2025imgedit}. Furthermore, we conduct a qualitative comparison against leading models on complex editing tasks to demonstrate practical utility.

\subsubsection{Benchmarks}

\paragraph{CEdit.} While numerous benchmarks exist for image editing, they often exhibit specific limitations in task coverage and granularity. For instance, GEdit-Bench, despite its popularity, lacks tasks involving reference image generation, structural modification, and viewpoint transformation. Similarly, ImgEdit-Bench offers a limited scope, KontextBench~\citep{batifol2025fluxkontext} suffers from coarse task granularity and low instruction diversity, and Reason-Edit~\citep{huang2024smartedit} prioritizes reasoning over conventional editing capabilities.

\begin{figure}[!tb]
\centering
    \includegraphics[width=1.0\linewidth]{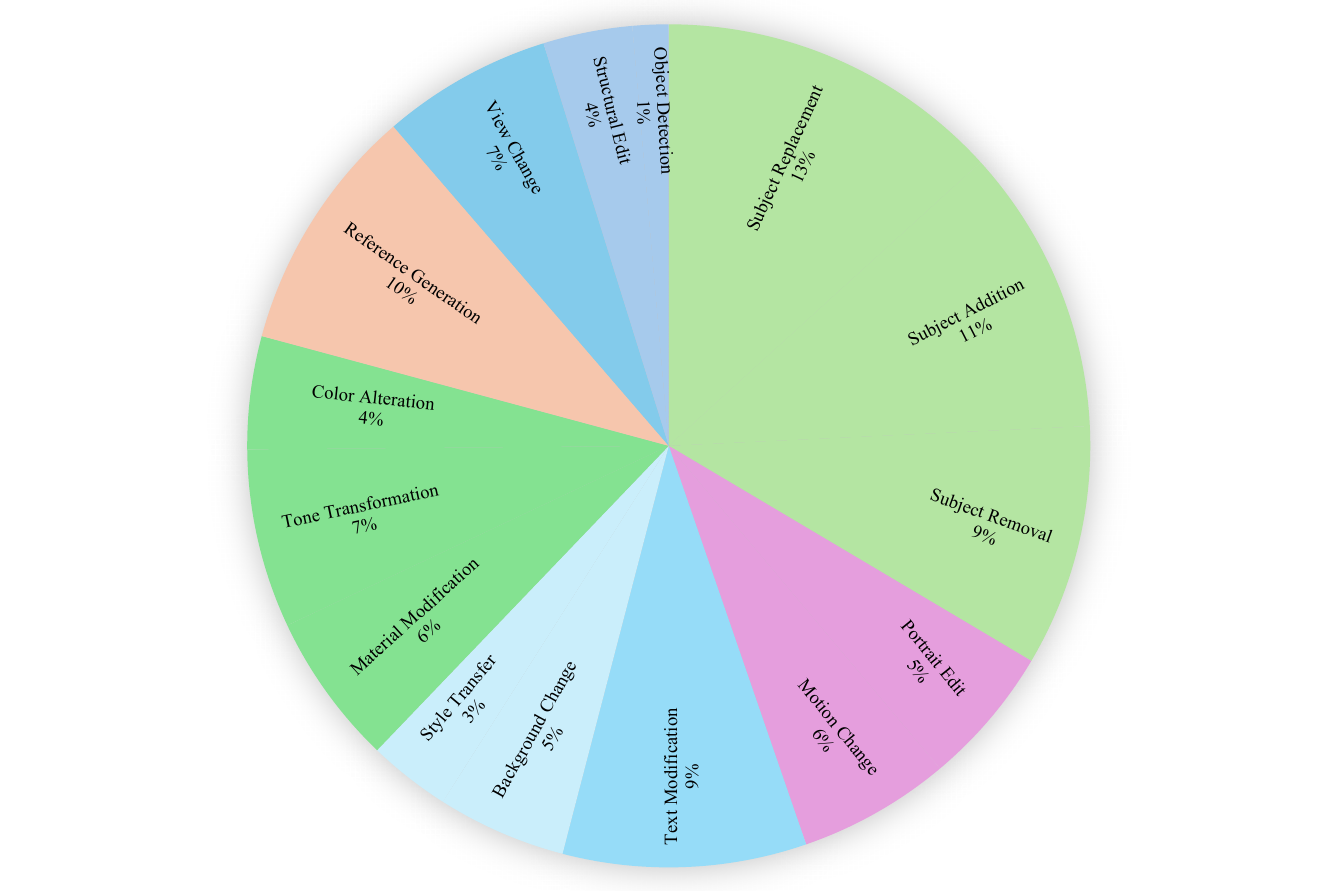}
\caption{\textbf{Task category distribution of CEdit-Bench.}}
\label{fig:exp_cedit}
\end{figure}

To address these gaps, we introduce CEdit-Bench, a comprehensive evaluation suite derived from the integration and expansion of these existing benchmarks. We curate new data to enrich task diversity, resulting in a robust dataset comprising 1,464 bilingual (Chinese and English) editing pairs across 15 fine-grained task categories, as illustrated in Fig.~\ref{fig:exp_cedit}. This establishes CEdit-Bench as a more holistic and rigorous evaluation standard.

We benchmark our model against FLUX.1 Kontext~\citep{batifol2025fluxkontext}, Step1X-Edit~\citep{liu2025step1x}, Qwen-Image-Edit~\citep{wu2025qwenimage}, Seedream 4.0~\citep{seedream2025seedream4}, and Nano Banana (Gemini-2.5-flash-image)\footnote{\url{https://aistudio.google.com/models/gemini-2-5-flash-image}} on CEdit-Bench. Following standard evaluation protocols, we employ Semantic Consistency (SQ), Perceptual Quality (PQ), and an Overall Score (O) as metrics, utilizing GPT-4o for automated evaluation. As shown in Table~\ref{tab:cedit_bench_results}, our model achieves SOTA performance among open-source models.

\begin{table}[!htb]
    \centering
    \caption{Performance comparison on CEdit-Bench.}
    \vspace{3pt}
    \renewcommand{\arraystretch}{1.25}
    \begin{tabular}{lcccccc}
        \toprule
        \multirow{2}{*}{\textbf{Model}} & \multicolumn{3}{c}{\textbf{CEdit-Bench-EN}$\uparrow$} & \multicolumn{3}{c}{\textbf{CEdit-Bench-CN}$\uparrow$} \\
        \cmidrule(lr){2-4} \cmidrule(lr){5-7}
        & \textbf{G\_SC} & \textbf{G\_PQ} & \textbf{G\_O} & \textbf{G\_SC} & \textbf{G\_PQ} & \textbf{G\_O} \\
        \midrule
        FLUX.1 Kontext [Pro]~\citep{batifol2025fluxkontext} &6.79 &7.80 &6.53 &1.15 &8.07 &1.43  \\
        GPT Image 1 [High]~\citep{gptimage} &\textbf{8.64} &\textbf{8.26} &\textbf{8.17} & \textbf{8.67}&\textbf{8.26} &\textbf{8.21}  \\
        Nano Banana~\citep{gemini25flash} & 7.51 & 8.17 & 7.20 & 7.67 & 8.21 & 7.36  \\
        Seedream 4.0~\citep{seedream2025seedream4} & 8.12 & 7.95 & 7.58 & 8.14 & 7.95 & 7.57\\
        \midrule
        FLUX.1 Kontext [Dev]~\citep{batifol2025fluxkontext} & 6.31 & 7.56 & 5.93 & 1.25 & 7.66 & 1.51 \\
        Step1X-Edit~\citep{liu2025step1x} & 6.68 & 7.36 & 6.25 & 6.88 & 7.28 & 6.35 \\
        Qwen-Image-Edit~\citep{wu2025qwenimage} & 8.07 & 7.84 & 7.52 & 8.03 & 7.78 & 7.46 \\
        Qwen-Image-Edit [2509]~\citep{wu2025qwenimage} & 8.04 & 7.79 & 7.48 &7.93 & 7.71 & 7.37 \\
        \textbf{LongCat-Image-Edit} & \textbf{8.27} & \textbf{7.88} & \textbf{7.67} & \textbf{8.25} & \textbf{7.85} & \textbf{7.65} \\
        \bottomrule
    \end{tabular}
    \label{tab:cedit_bench_results}
\end{table}

\paragraph{GEdit.} To benchmark against established standards, we evaluate our model on GEdit-Bench, comparing it with both popular open-source and proprietary products. As reported in Table~\ref{tab:gedit_bench_results}, our model achieves top-tier performance, demonstrating superior instruction-following capabilities in both Chinese and English.

\begin{table}[!htb]
    \centering
    \caption{Performance comparison on GEdit-Bench.}
    \vspace{3pt}
    \renewcommand{\arraystretch}{1.25}
    \begin{tabular}{lcccccc}
        \toprule
        \multirow{2}{*}{\textbf{Model}} & \multicolumn{3}{c}{\textbf{GEdit-Bench-EN}$\uparrow$} & \multicolumn{3}{c}{\textbf{GEdit-Bench-CN}$\uparrow$} \\
        \cmidrule(lr){2-4} \cmidrule(lr){5-7}
        & \textbf{G\_SC} & \textbf{G\_PQ} & \textbf{G\_O} & \textbf{G\_SC} & \textbf{G\_PQ} & \textbf{G\_O} \\
        \midrule
        Gemini 2.0~\citep{GoogleDeepMindGemini2025} & 6.73 & 6.61 & 6.32 & 5.43 & 6.78 & 5.36 \\
        FLUX.1 Kontext [Pro]~\citep{batifol2025fluxkontext} & 7.02 & 7.60 & 6.56 & 1.11 & 7.36 & 1.23 \\
        GPT Image 1 [High]~\citep{gptimage} & 7.85 & 7.62 & 7.53 & 7.67 & 7.56 & 7.30 \\
        Nano Banana~\citep{gemini25flash} &7.86  &\textbf{8.33}  &7.54 &7.51  & \textbf{8.31} & 7.25 \\
        Seedream 4.0~\citep{seedream2025seedream4}  & \textbf{8.24} & 8.08 & \textbf{7.68} &\textbf{8.19}  &8.14  &\textbf{7.71}  \\
        \midrule
        InstructPix2Pix \citep{brooks2023instructpix2pix} & 3.58 & 5.49 & 3.60 & - & - & - \\
        AnyEdit~\citep{yu2025anyedit} & 3.18 & 5.82 & 3.21 & - & - & - \\
        MagicBrush~\citep{zhang2023magicbrush} & 4.68 & 5.66 & 4.52 & - & - & - \\
        UniWorld-v1~\citep{lin2025uniworld} & 4.93 & 7.43 & 4.85 & - & - & - \\
        OmniGen~\citep{xiao2025omnigen} & 5.96 & 5.89 & 5.06 & - & - & - \\
        OmniGen2~\citep{wu2025omnigen2} & 7.16 & 6.77 & 6.41 & - & - & - \\
        FLUX.1 Kontext [Dev]~\citep{batifol2025fluxkontext} & 6.52 & 7.38 & 6.00 & - & - & - \\
        BAGEL~\citep{deng2025emerging} & 7.36 & 6.83 & 6.52 & 7.34 & 6.85 & 6.50 \\
        Step1X-Edit~\citep{liu2025step1x} & 7.66 & 7.35 & 6.97 & 7.20 & 6.87 & 6.86 \\
        Qwen-Image-Edit~\citep{wu2025qwenimage}& 8.00 & 7.86 & 7.56 & 7.82 & 7.79 & 7.52 \\
        Qwen-Image-Edit [2509]~\citep{wu2025qwenimage} & 8.15 & 7.86 & 7.54 & 8.05 & 7.88 & 7.49 \\
        \textbf{LongCat-Image-Edit} & \textbf{8.18} & \textbf{8.00} & \textbf{7.64} & \textbf{8.08} & \textbf{7.99} &\textbf{7.60} \\
        \bottomrule
    \end{tabular}
    \label{tab:gedit_bench_results}
\end{table}

\paragraph{ImgEdit.} Serving as a complement to GEdit, ImgEdit-Bench evaluates models with a focus on instruction adherence, editing quality, and detail preservation. We compare our model against competitive baselines using the official metrics provided by the benchmark. The results in Table~\ref{tab:imgedit-benchmark-results} indicate that our model outperforms all competitors, further validating its comprehensive capabilities across diverse evaluation dimensions.

\begin{table}[!htb]
    \centering
    \caption{Performance comparison on ImgEdit-Bench.}
    \vspace{3pt}
    \renewcommand{\arraystretch}{1.25}
    \resizebox{0.99\textwidth}{!}{
    \begin{tabular}{lcccccccccc}
        \toprule
        \textbf{Model} & \textbf{Add} & \textbf{Adjust} & \textbf{Extract} & \textbf{Replace} & \textbf{Remove} & \textbf{Background} & \textbf{Style} & \textbf{Hybrid} & \textbf{Action} & \textbf{Overall}$\uparrow$ \\
        \midrule
        FLUX.1 Kontext [Pro]~\citep{batifol2025fluxkontext}  & 4.25 & 4.15 & 2.35 & 4.56 & 3.57 & 4.26 & 4.57 & 3.68 & 4.63 & 4.00 \\
        GPT Image 1 [High]~\citep{gptimage} & \textbf{4.61} & 4.33 & 2.90 & 4.35 & 3.66 & 
        \textbf{4.57} & \textbf{4.93} & 3.96 & \textbf{4.89} & 4.20 \\
        Nano Banana~\citep{gemini25flash} &4.50  &\textbf{4.47} &\textbf{3.75} &\textbf{4.64} &\textbf{4.51} &4.44 &4.14 &\textbf{4.03} &4.65 &\textbf{4.35} \\
        Seedream4.0~\citep{seedream2025seedream4} &4.52  &4.41 &2.93 &4.56 &4.44 &4.30 &4.76 &3.33 &4.36 &4.18 \\
        \midrule
        MagicBrush~\citep{zhang2023magicbrush}  & 2.84 & 1.58 & 1.51 & 1.97 & 1.58 & 1.75 & 2.38 & 1.62 & 1.22 & 1.90 \\
        InstructPix2Pix~\citep{brooks2023instructpix2pix} & 2.45 & 1.83 & 1.44 & 2.01 & 1.50 & 1.44 & 3.55 & 1.20 & 1.46 & 1.88 \\
        AnyEdit~\citep{yu2025anyedit}   & 3.18 & 2.95 & 1.88 & 2.47 & 2.23 & 2.24 & 2.85 & 1.56 & 2.65 & 2.45 \\
        UltraEdit~\citep{zhao2024ultraedit}   & 3.44 & 2.81 & 2.13 & 2.96 & 1.45 & 2.83 & 3.76 & 1.91 & 2.98 & 2.70 \\
        OmniGen~\citep{xiao2025omnigen}   & 3.47 & 3.04 & 1.71 & 2.94 & 2.43 & 3.21 & 4.19 & 2.24 & 3.38 & 2.96 \\
        ICEdIt~\citep{zhang2025context} & 3.58 & 3.39 & 1.73 & 3.15 & 2.93 & 3.08 & 3.84 & 2.04 & 3.68 & 3.05 \\
        Step1X-Edit~\citep{liu2025step1x} & 3.88 & 3.14 & 1.76 & 3.40 & 2.41 & 3.16 & 4.63 & 2.64 & 2.52 & 3.06 \\
        BAGEL~\citep{deng2025emerging} & 3.56 & 3.31 & 1.70 & 3.30 & 2.62 & 3.24 & 4.49 & 2.38 & 4.17 & 3.20 \\
        UniWorld-V1~\citep{lin2025uniworld} & 3.82 & 3.64 & 2.27 & 3.47 & 3.24 & 2.99 & 4.21 & 2.96 & 2.74 & 3.26 \\
        OmniGen2~\citep{wu2025omnigen2}  & 3.57 & 3.06 & 1.77 & 3.74 & 3.20 & 3.57 & 4.81 & 2.52 & 4.68 & 3.44 \\
        FLUX.1 Kontext [Dev]~\citep{batifol2025fluxkontext} & 4.12 & 3.80 & 2.04 & 4.22 & 3.09 & 3.97 & 4.51 & 3.35 & 4.25 & 3.71 \\
        Qwen-Image-Edit~\citep{wu2025qwenimage}     & 4.38 & 4.16 & 3.43 & 4.66 & 4.14 & 4.38 & 4.81 & 3.82 & 4.69 & 4.27 \\
        Qwen-Image-Edit [2509]~\citep{wu2025qwenimage}  & 4.32 & 4.36 & \textbf{4.04} & 4.64 & 4.52 & 4.37 & 4.84 & 3.39 & 4.71 & 4.35 \\
        \textbf{LongCat-Image-Edit}      & \textbf{4.51} & \textbf{4.57} & 3.93 & \textbf{4.76} & \textbf{4.60} & \textbf{4.49} & \textbf{4.85} & \textbf{4.01} & \textbf{4.74} & \textbf{4.50} \\
        \bottomrule
    \end{tabular}}
    \label{tab:imgedit-benchmark-results}
\end{table}

\subsubsection{Human Evaluation}
To benchmark our model against SOTA open-source models and leading commercial products, we conduct a Side-by-Side (SBS) human evaluation. This assessment focuses on two primary dimensions: comprehensive quality and consistency.

\begin{itemize}
    \item \textbf{Comprehensive Quality.} This metric evaluates the overall performance of image editing across multiple aspects, including instruction adherence, visual plausibility, aesthetics, and the consistency between original and edited images. Annotators provide a holistic judgment by categorizing the result as a \textit{Win}, \textit{Tie}, or \textit{Loss}.
    
    \item \textbf{Consistency.} We conduct a dedicated evaluation for this dimension, distinct from the comprehensive score, to emphasize its critical role in multi-turn editing. This metric specifically scrutinizes whether attributes in non-edited regions, such as layout, texture, color tone, and subject identity, remain invariant unless targeted by the instruction.
\end{itemize}

\paragraph{\textbf{Evaluation Dataset.}} We curate a diverse dataset comprising approximately 400 samples tailored for black-box evaluation. The dataset covers a broad spectrum of difficulty levels and includes various editing tasks, such as global editing, local editing, text modification, and reference-guided editing.

\begin{figure}[!tb]
    \centering
    \includegraphics[width=1.0\linewidth]{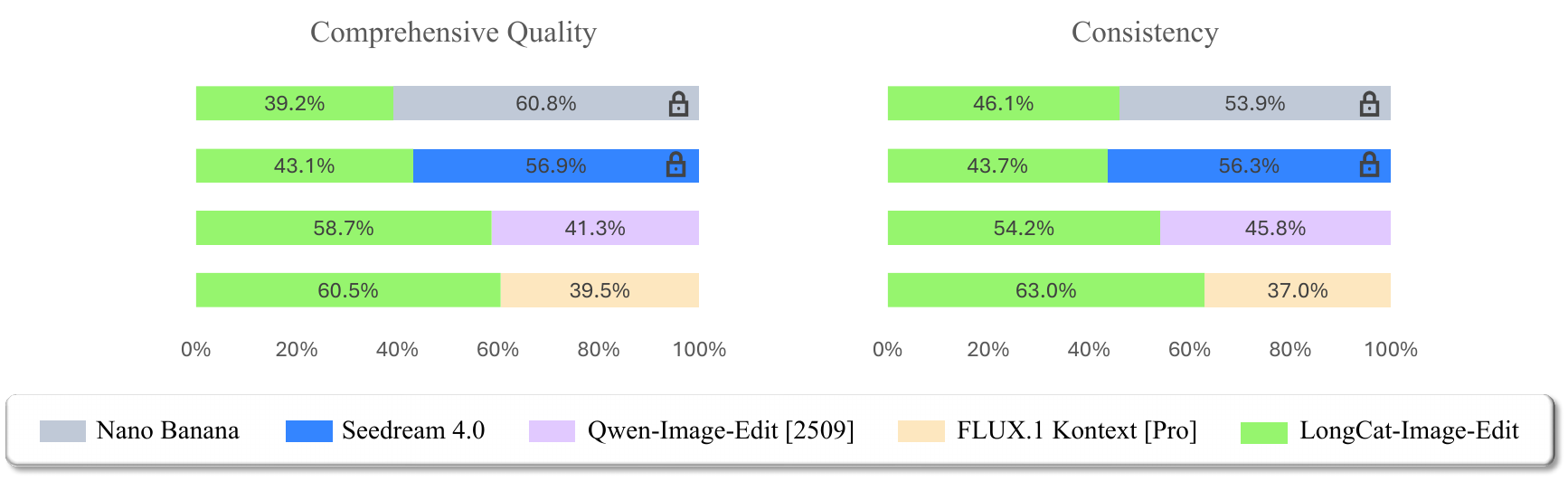}
    \caption{\textbf{Comparison of human evaluation win rates between LongCat-Image-Edit and competing methods.}}
    \label{fig:edit_human_winrate}
\end{figure}

\paragraph{\textbf{Result Analysis.}} We calculate the win rate using the formula: $(\text{\#Win} + 0.5 \times \text{\#Tie}) / \text{\#Total}$. As illustrated in Fig.~\ref{fig:edit_human_winrate}, LongCat-Image-Edit outperforms Qwen-Image-Edit [2509] and FLUX.1 Kontext [Pro] in both comprehensive quality and consistency. However, a performance gap remains when compared to commercial systems such as Nano Banana and Seedream 4.0.

\subsubsection{Qualitative Results}
To comprehensively evaluate our model's versatility, we conduct qualitative comparisons against leading instruction-based image editing baselines. We begin by highlighting the model's performance in two distinct, high-demand real-world scenarios: \textbf{Multi-turn Editing} and \textbf{Portrait and Human-Centric Editing}. These tasks are selected for their prevalence in practical applications and the rigorous requirements they impose on editing precision.

\begin{figure}[h!]
    \centering
    \includegraphics[trim={0cm 0cm 0.2cm 0cm}, clip, width=1.0\textwidth]{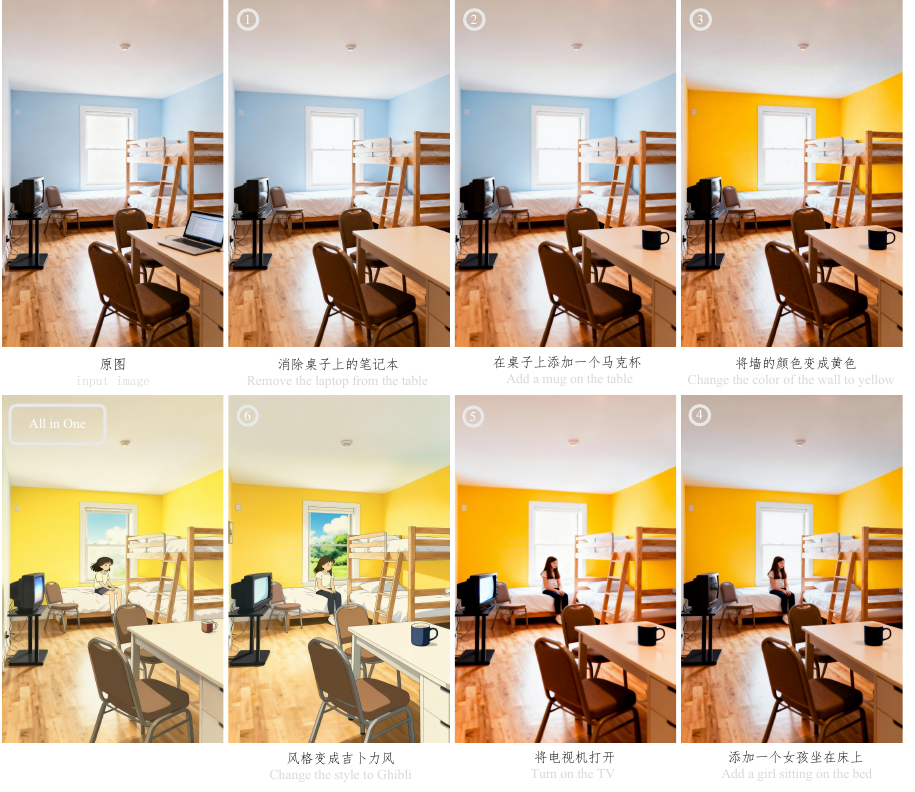} 
    \caption{\textbf{Visual comparison of multi-turn editing versus one-shot composite editing.} The numbered sequence (\textcircled{1}--\textcircled{6}) illustrates the progressive results of multi-turn editing. In contrast, the ``All in One'' image (bottom-left) demonstrates the outcome of a single complex instruction containing all six operations: \textit{Remove the laptop, add a mug, change the wall to yellow, add a girl sitting on the bed, turn on the TV, and change the style to Ghibli. }}
    \label{fig:edit_chain}
\end{figure}

\paragraph{Multi-turn Editing.} Sequential editing imposes stringent demands on a model's ability to preserve visual consistency across iterative steps. We evaluate our model on representative editing chains and further challenge it with \textit{compound instructions}—where multiple edits are requested in a single prompt. As illustrated in Fig.~\ref{fig:edit_chain}, our model maintains exceptional semantic and structural consistency throughout the entire editing sequence. Remarkably, even when processing a complex prompt containing six distinct operations, the model executes all directives accurately. The result aligns closely with the sequential output, underscoring its robust capability in handling both fine-grained iterative updates and complex composite tasks.

\begin{CJK}{UTF8}{gbsn}
\end{CJK}

\begin{figure}[!htb]
    \centering
    \includegraphics[width=1.0\textwidth]{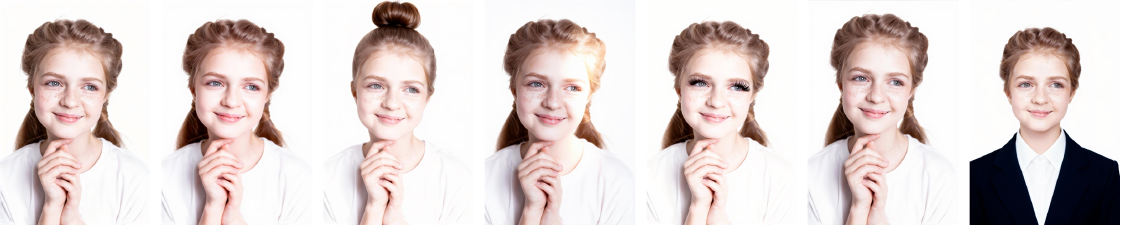}
    \caption{\textbf{Demonstration of fine-grained portrait editing.} From left to right: The original input image, followed by results for blemish removal, hairstyle modification, lighting adjustment, eyelash addition, face slimming, and ID photo generation. The results highlight the model's precision in manipulating specific facial attributes while preserving the subject's identity.}
    \label{fig:edit_human_beauty}
    \vspace{20pt}
\end{figure}

\begin{figure}[!h]
    \centering
    \includegraphics[width=1.0\textwidth]{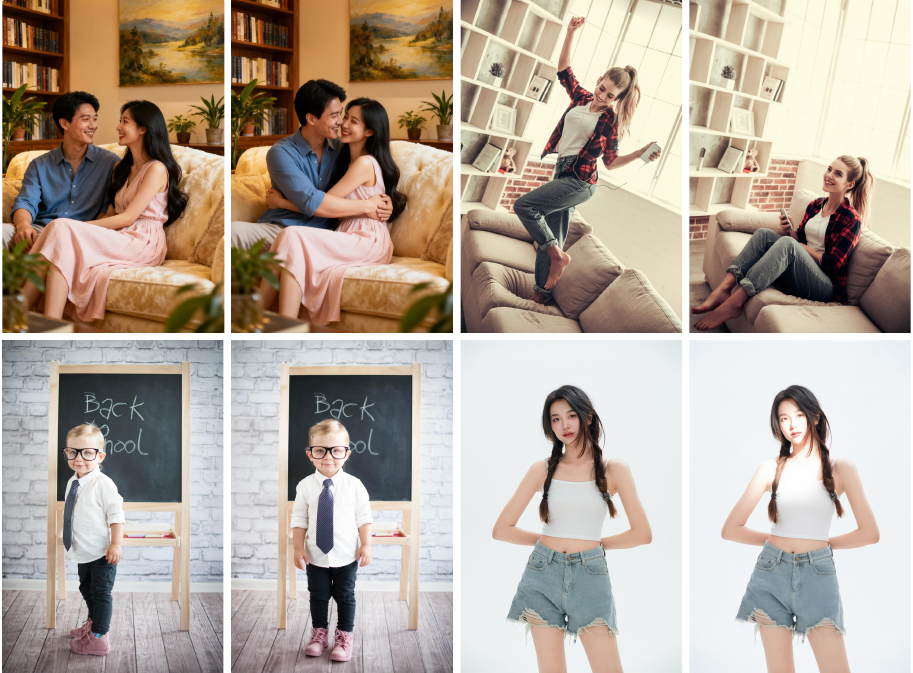} 
    \caption{\textbf{Qualitative results on Human-centric Editing.} The figure displays pairs of input (left) and edited (right) images across three dimensions: \textbf{Pose \& Interaction} (top row), involving complex interaction synthesis (\textit{e.g.}, hugging) and large-scale body pose alteration; \textbf{Viewpoint} (bottom-left), transforming a subject from side view to front view; and \textbf{Lighting} (bottom-right), simulating directional illumination. Note the preservation of background details and subject identity despite significant structural changes.}
    \label{fig:edit_human_pose}
    \vspace{20pt}
\end{figure}

\paragraph{Portrait and Human-Centric Editing.}  Fig.~\ref{fig:edit_human_beauty} validates the model's precision in fine-grained portrait editing, confirming its ability to accurately execute diverse facial attribute modifications while preserving identity. Expanding beyond facial details, Fig.~\ref{fig:edit_human_pose} demonstrates robust performance in structural body editing. The model successfully handles complex challenges ranging from viewpoint transformation to multi-person interaction synthesis. Collectively, these results highlight the model's significant practical utility, indicating its potential for integration into mobile photography pipelines for high-quality post-processing.

\begin{figure}[!hp]
    \centering
    \includegraphics[width=0.9\textwidth]{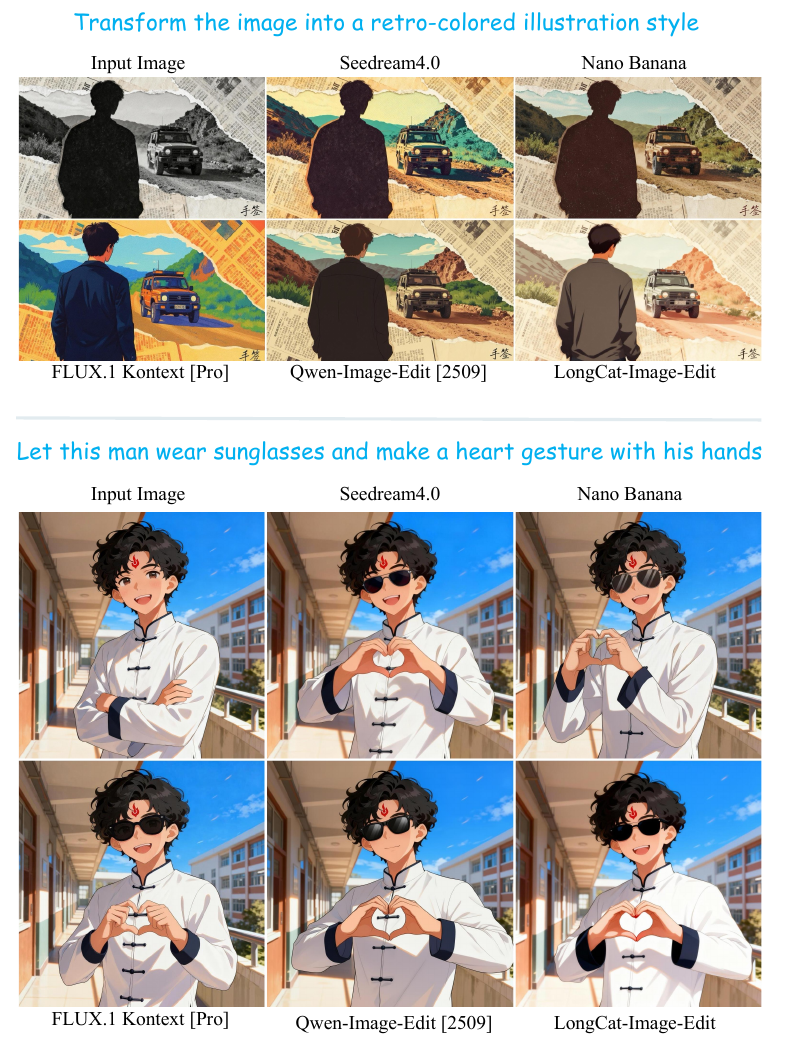}
    \caption{\textbf{Qualitative comparison on Style Transfer and Attribute Editing.} The upper panel demonstrates the transformation of a photorealistic scene into a retro-colored illustration style. The lower panel illustrates a complex instruction involving both accessory addition (sunglasses) and hand pose modification (heart gesture), highlighting our model's ability to preserve facial identity while executing significant structural changes.}
\label{fig:edit_general_compare1}
\end{figure}

\begin{figure}[!hp]
    \centering
    \includegraphics[width=0.95\textwidth]{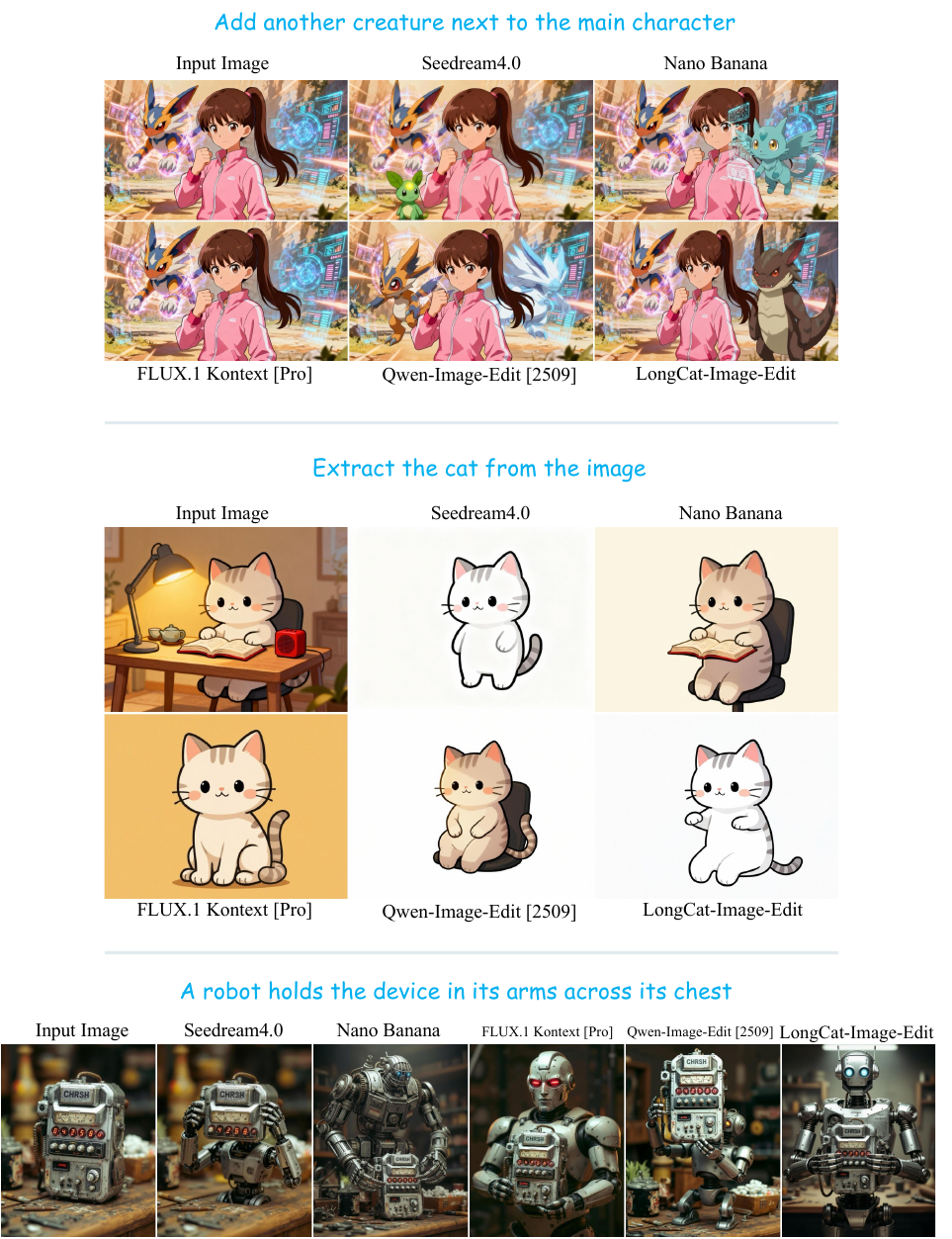}
    \caption{\textbf{Qualitative comparison on Object-centric Editing.} We evaluate the performance across three distinct scenarios: \textbf{Object Insertion}   (top), where an additional creature is added while maintaining scene consistency; \textbf{Subject Extraction} (middle), isolating the foreground subject (the cat) from a complex background; \textbf{Object-Preserved Generation} (bottom), where the reference object (the device) is seamlessly integrated into a new context (held by a robot).}
\label{fig:edit_general_compare2}
\end{figure}

\begin{figure}[!hp]
    \centering
    \includegraphics[width=0.89\textwidth]{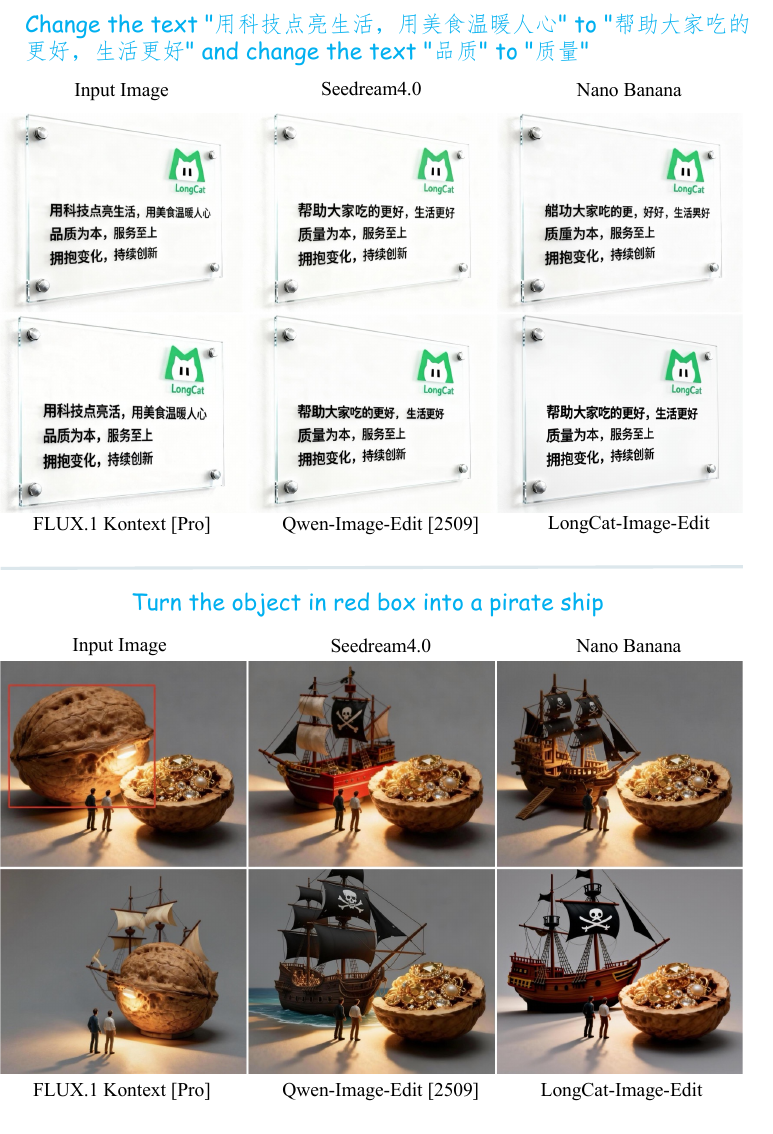}
    \vspace{-7pt}
    \caption{\textbf{Qualitative comparison on Scene Text Editing and Region-Controlled Editing.}}
\label{fig:edit_general_compare3}
\end{figure}

\begin{figure}[!hp]
    \centering
    \includegraphics[width=0.95\textwidth]{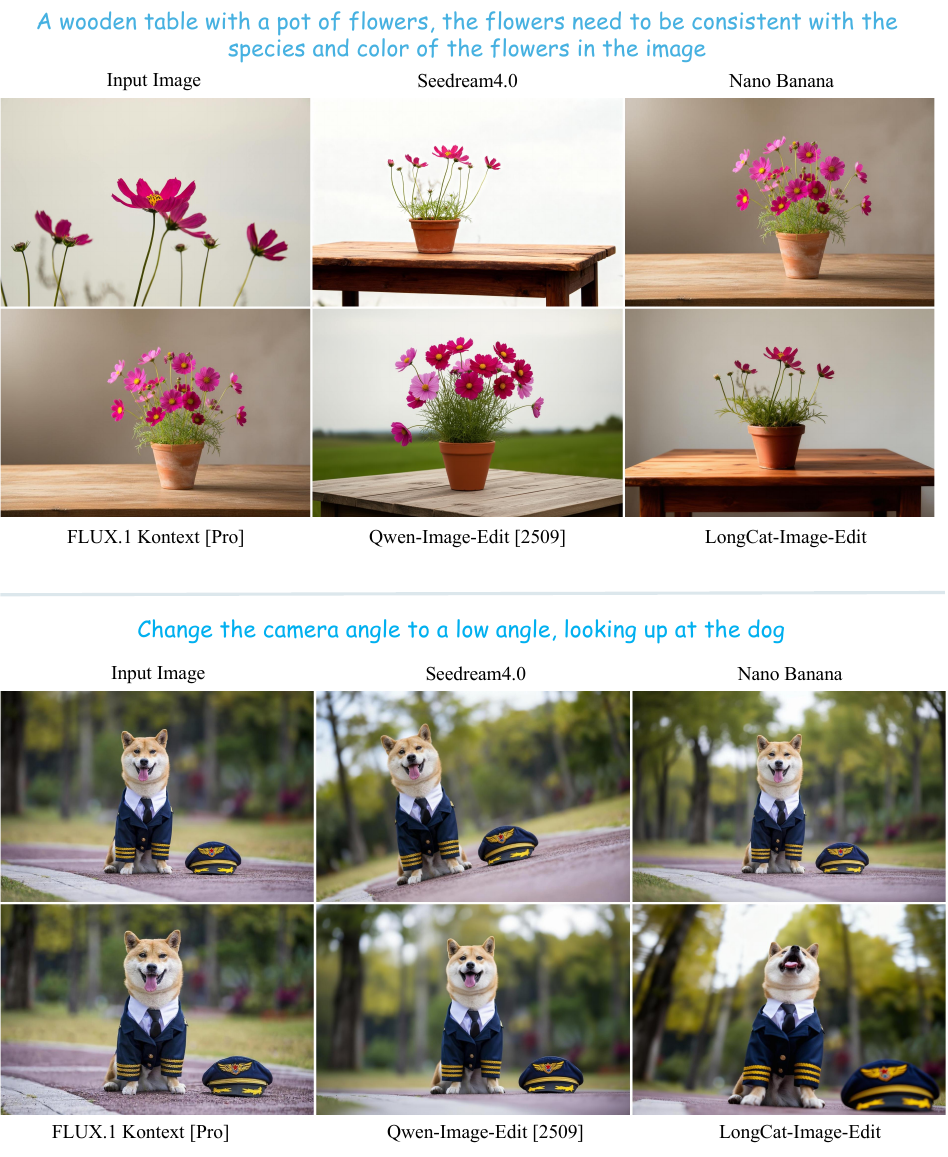}
    \caption{\textbf{Qualitative comparison on Camera Control and Viewpoint Transformation.} The upper panel shows \textbf{camera distance adjustment} (zooming out), where the view is expanded to place the flowers into a pot on a table. The lower panel displays \textbf{camera angle modification}, transitioning the view to a low-angle perspective looking up at the dog. }
\label{fig:edit_general_compare4}
\end{figure}

\paragraph{General Image Editing Capabilities.} We conduct a comprehensive qualitative evaluation against SeedDream 4.0, Nano Banana, Qwen-Image-Edit [2509], and FLUX.1 Kontext [Pro]. The assessment covers a broad spectrum of editing dimensions: object manipulation (addition, removal, extraction), attribute modification, viewpoint transformation, scene text editing, and reference-guided generation. As illustrated in Fig.~\ref{fig:edit_general_compare1},~\ref{fig:edit_general_compare2},~\ref{fig:edit_general_compare3},~\ref{fig:edit_general_compare4}, our model consistently outperforms Qwen-Image-Edit [2509] and FLUX.1 Kontext [Pro] across all dimensions. Notably, in certain challenging scenarios, it also achieves superior results compared to the commercial counterparts.

\section{Conclusion}
\label{sec:conclusion}

In this work, we present \textbf{LongCat-Image}, a 6B-parameter diffusion framework that challenges the prevailing reliance on brute-force scaling by demonstrating that exceptional performance can be achieved through efficient architectural design and refined training methodologies. By integrating a hybrid MM-DiT architecture with a unified multimodal context encoder, our model establishes an optimal equilibrium between high-fidelity generation and inference efficiency, effectively surpassing the generation quality of numerous open-source models with significantly larger parameters.
Across specific domains, LongCat-Image delivers exceptional results. In text-to-image generation, our strategic data curation enables photorealism and Chinese text rendering capabilities that compete with top-tier proprietary systems. In the realm of image editing, our model sets a new benchmark for the open-source community. Supported by rigorous data filtering and a robust training paradigm, it achieves state-of-the-art performance, exhibiting both precise instruction following and superior visual consistency that significantly outperform existing alternatives.
Finally, we distinguish our contribution by democratizing the entire research lifecycle. By open-sourcing not only the final model but also intermediate checkpoints and the complete training codebase, we aim to lower the barriers to entry and foster a more transparent, accessible, and collaborative ecosystem for future research.
\clearpage
\section{Contributions and Acknowledgments}

Contributors are defined as individuals who undertook primary responsibilities in data curation, model design, model training, and relative infrastructures throughout the LongCat-Image complete development cycle. Acknowledgment include those who are working part-time on tasks such as data collection, annotation, model evaluation, and technical discussions. All people are cataloged \textbf{alphabetically by first name}. 
Names with a dagger ($^\dagger$) are the project leader and sponsors, and names with an asterisk ($^\ast$) are former team members.

\vspace{10pt}

\noindent\textbf{Contributors:}
\begin{multicols}{4}
\noindent
\raggedright
\begin{spacing}{1.2}
Hanghang Ma$^\ast$ \\
Haoxian Tan \\
Jiale Huang \\
Jie Hu$^\dagger$ \\
Junqiang Wu \\
Jun-Yan He \\
Lishuai Gao \\
Songlin Xiao \\
Xiaoming Wei$^\dagger$ \\
Xiaoqi Ma$^\ast$ \\
Xunliang Cai$^\dagger$ \\
Yayong Guan \\
\end{spacing}
\end{multicols}

\vspace{5pt}
\noindent\textbf{Acknowledgments:}
\begin{multicols}{4}
\noindent
\raggedright
\begin{spacing}{1.2}
Bingcan Wang \\ Cong Wei \\ Dengsheng Chen$^\ast$ \\ Fei Peng \\ Fengjiao Chen \\ Hao Lu \\ Jia Wang \\ Jiajun Liu \\  Kaiwen Wang$^\ast$ \\ Lingfeng Tan$^\ast$ \\ Liya Ma \\ Man Gao \\ Shengxi Li \\ Tianye Dai \\ Tiezhu Yue \\ Wei Wang \\ Xiaopeng Sun$^\ast$ \\ Xiaoyu Li  \\ Yanbing Zeng \\ Yingsen Zeng$^\ast$ \\ Yuchen Tang \\ Zizhe Zhao
\end{spacing}
\end{multicols}

\bibliographystyle{unsrtnat}
\bibliography{references}

\end{document}